\documentclass[10pt,twocolumn,letterpaper]{article}

\usepackage{cvpr}
\usepackage{times}
\usepackage{paralist}
\usepackage{epsfig}
\usepackage{graphicx}
\usepackage{amsmath}
\usepackage{amsthm}
\usepackage{amssymb}
\usepackage{enumitem}
\usepackage{caption}
\usepackage{subcaption}

\usepackage[pagebackref=true,breaklinks=true,letterpaper=true,colorlinks,bookmarks=false]{hyperref}

\cvprfinalcopy 

\def\cvprPaperID{174} 

\ifcvprfinal\fi

\newcommand{\figref}[1]{Fig.~\ref{fig:#1}}
\newcommand{\secref}[1]{Sec.~\ref{sec:#1}}
\newcommand{\equref}[1]{\eqref{eq:#1}}
\newcommand{\tabref}[1]{Tab.~\ref{tab:#1}}

\newcommand{\topt}{\mathsf{T}}

\newcommand{\myB}{\mathbf{B}}
\newcommand{\myp}{\mathbf{p}}
\newcommand{\myl}{\mathbf{l}}
\newcommand{\myL}{\mathbf{L}}
\newcommand{\myh}{\mathbf{h}}

\newcommand{\myz}{\mathbf{z}}
\newcommand{\myang}{\Theta}
\newcommand{\given}{\vert}

\graphicspath{
  {figs/}
  {figs/cover/}
  {figs/analysis/}
  {figs/montage/}
  {figs/sampling/}
  {figs/deepnet/}
  {figs/geometry/}
  {figs/failures/}
  {figs/vp/}
  {figs/pipeline/}
  {figs/pdf/}}

\begin{document}

\title{Detecting Vanishing Points using Global Image \\
  Context in a Non-Manhattan World}

\author{
  \begin{minipage}{\linewidth}
    \centering
    \begin{minipage}{1.5in}
      \centering
      Menghua Zhai\\
      {\tt\small ted@cs.uky.edu}
    \end{minipage}
    \begin{minipage}{1.5in}
      \centering
      Scott Workman\\
      {\tt\small scott@cs.uky.edu}
    \end{minipage}
    \begin{minipage}{1.5in}
      \centering
      Nathan Jacobs\\
      {\tt\small jacobs@cs.uky.edu}
    \end{minipage}
    \\[.2cm]
    \begin{minipage}{4in}
      \centering
      Computer Science, University of Kentucky \\ 
    \end{minipage}
  \end{minipage}
}

\maketitle

\begin{abstract}

  We propose a novel method for detecting horizontal vanishing points
  and the zenith vanishing point in man-made environments. The
  dominant trend in existing methods is to first find candidate
  vanishing points, then remove outliers by enforcing mutual
  orthogonality. Our method reverses this process: we propose a set of
  horizon line candidates and score each based on the vanishing points
  it contains. A key element of our approach is the use of global
  image context, extracted with a deep convolutional network, to
  constrain the set of candidates under consideration. Our method does
  not make a Manhattan-world assumption and can operate effectively on
  scenes with only a single horizontal vanishing point.  We evaluate
  our approach on three benchmark datasets and achieve
  state-of-the-art performance on each. In addition, our approach is
  significantly faster than the previous best method.

\end{abstract}




\section{Introduction}

Automatic vanishing point (VP) and horizon line detection are two of
the most fundamental problems in geometric computer
vision~\cite{barnard1983interpreting,magee1984determining}. Knowledge
of these quantities is the foundation for many higher level
tasks, including image mensuration~\cite{criminisi2000single},
facade detection~\cite{liulocal2014},
geolocalization~\cite{baatz2010handling,workman2014rainbow}, and
camera
calibration~\cite{autorecovery2000,grammatikopoulos2007automatic,
jacobs13cloudcalibration,videocampass2002}. Recent work in this area
\cite{global2013,wildenauer2012,kitware2013} has explored novel
problem formulations that significantly increase robustness to noise.

\begin{figure}
  \centering
  \includegraphics[width=.48\linewidth]{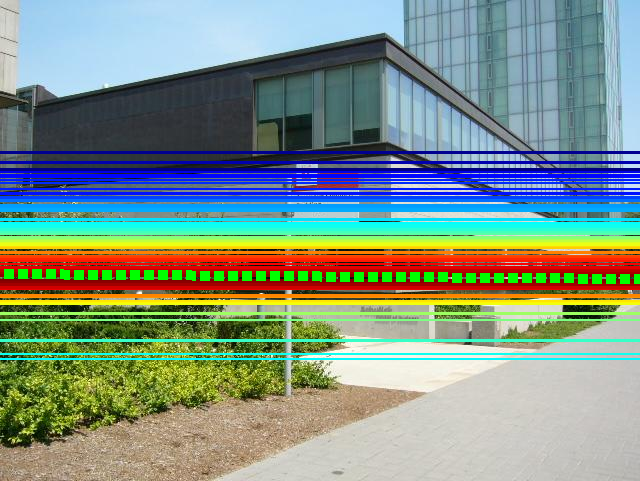}
  \includegraphics[width=.48\linewidth]{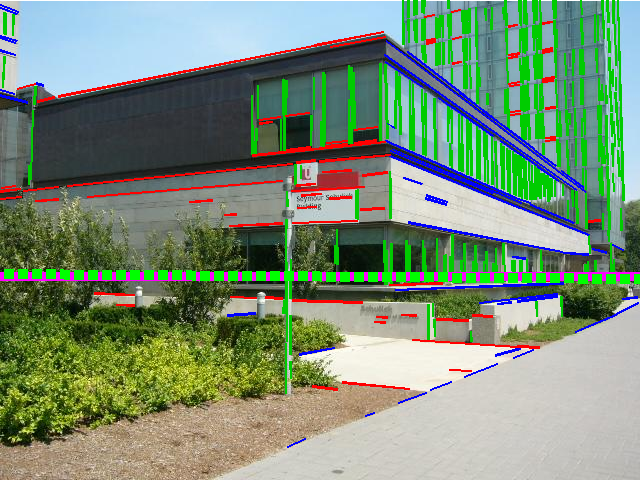}

  \caption{An example result of our method. (left) Horizon line
    candidates, colored by their scores (red means high score), and
    the true horizon line (green dash). (right) The horizon line
    (magenta) estimated by our algorithm is very close to the true
    horizon line (green dash).  Line segments are color coded based on
    the most consistent detected vanishing point.}

  \label{fig:cover}
\end{figure}

A vanishing point results from the intersection of projections of a
set of parallel lines in the world. In man-made environments, such
sets of lines are often caused by the edges of buildings, roads, and
signs. VPs can typically be classified as either vertical, there is
one such VP, and horizontal, there are often many such VPs.  Given a
set of horizontal VPs, there are numerous methods to estimate the
horizon line.  Therefore, previous approaches to this problem focus on
first detecting the vanishing points, which is a challenging problem
in many images due to line segment intersections that are not true
VPs.

Our approach is to propose candidate horizon lines, score them, and
keep the best (\figref{cover}). We use a deep convolutional neural network to extract
global image context and guide the generation of a set of horizon line
candidates. For each candidate, we identify vanishing points by
solving a discrete-continuous optimization problem.  The final score for
each candidate line is based on the consistency of the lines in the
image with the selected vanishing points. 

This seemingly simple shift in approach leads to the need for novel
algorithms and has excellent performance. We evaluated the proposed
approach on two standard benchmark datasets, the Eurasian Cities
Dataset~\cite{geoparser2010} and the York Urban
Dataset~\cite{edgebased2008}. To our knowledge, our approach has the
current best performance on both datasets. To evaluate our algorithm
further, we also compare with the previous state-of-the-art method
(Lezama et al.~\cite{alignment2014}) on a recently introduced
dataset~\cite{authors2016deephorizon}; the results shows that our
method is more accurate and much faster.

\begin{figure*}
  \centering
  \includegraphics[width=0.95\linewidth]{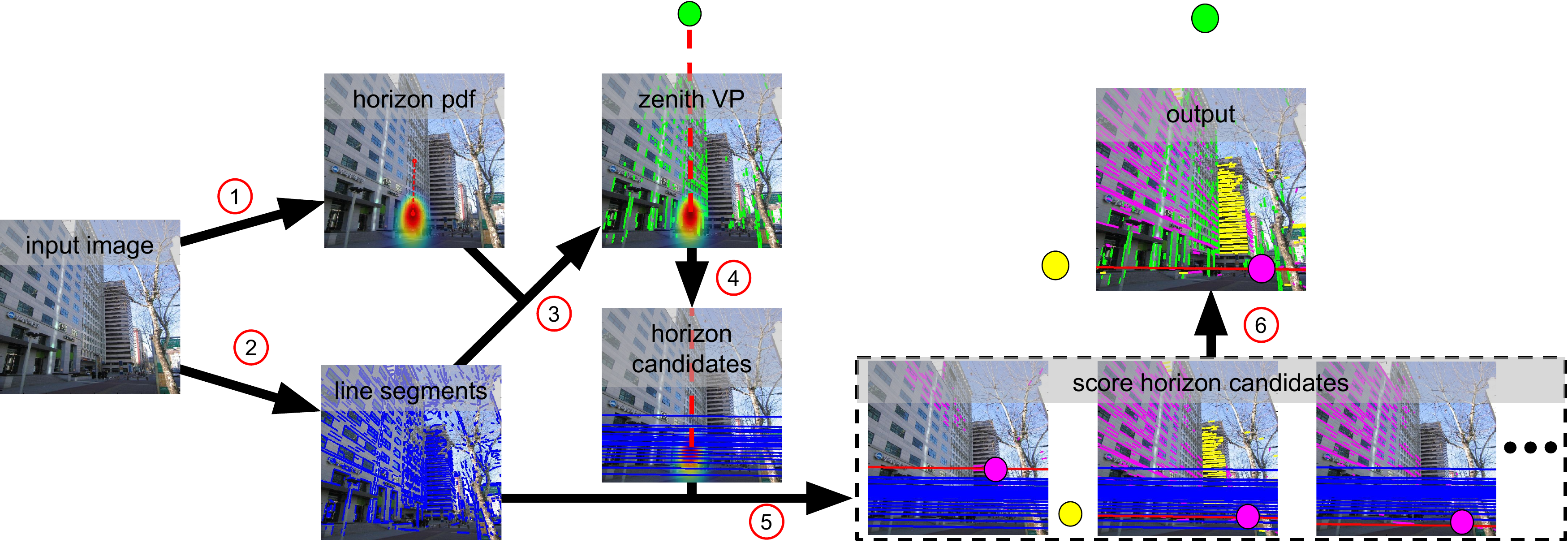}
  \caption{Algorithm overview: 1) use global image context to estimate
    a prior over horizon lines (\secref{deeplearning}); 2) extract
    line segments; 3) identify the zenith VP
    (\secref{zenithdetection}); 4) sample horizon line candidates
    consistent with the zenith VP (\secref{horizondetection}); 5) find
    VPs on horizon line candidates (\secref{horizondetection}); and 6)
    select the best horizon line based on the VPs it contains
    (\secref{horizonscoring}).}
  \label{fig:pipeline}
\end{figure*}


The main contributions of this work
are: 1) a novel method for horizon line/vanishing point detection,
which uses global image context to guide precise geometric analysis;
2) a strategy for quickly extracting this context, in the form of
constraints on possible horizon lines, using a deep convolutional
neural network; 3) a discrete-continuous method for scoring horizon
line candidates; and 4) an evaluation of the proposed approach on
three benchmark datasets, which highlights that our method is both
fast and accurate.

\subsection{Related Work}
\label{sec:relatedwork}

Vanishing points and the horizon line provide a strong
characterization of geometric scene structure and as such have been
intensely studied for
decades~\cite{barnard1983interpreting,magee1984determining}.  For
example, Hoiem et al.~\cite{hoiem2008putting} show how the horizon
line improves the accuracy of object detection.  A wide variety of
methods have been introduced to estimate these quantities.  We provide
a brief overview of the main approaches, refer
to~\cite{szeliski2010computer} for a comprehensive review. 

Two distinct categories of methods exist, distinguished by the
features they use. The first group of methods~\cite{geoparser2010,
manhattanbayesian1999, edgebased2008,atlanta2004} operate directly
on lower-level features, such as edge pixels or image gradients.  The
second group of methods~\cite{nopriori2003, edgebased2008,
alignment2014, houghvp1994, selfsketch2012, wildenauer2012,
kitware2013} build on top of the closely related problem of line
segment detection.  Our work is most closely related to the latter
category, so we focus our discussion towards them.

The dominant approach to vanishing point detection from line segments
is to cluster the line segments that pass through the same location.
Various methods of clustering have been explored, including
RANSAC~\cite{ransac1981}, J-linkage~\cite{tardif2009}, and the Hough
transform~\cite{hough1959}. Once the line segments have been
clustered, vanishing points can be estimated using one of many
refinement procedures~\cite{alignment2014,atlanta2004,tardif2009,
wildenauer2012,kitware2013}. 

These procedures typically minimize a
nonlinear objective function. An important distinction between such
methods is the choice of point and line representation and error
metric. Collins and Weiss~\cite{unitsphere1990} formulate vanishing
point detection as a statistical estimation problem on the Gaussian
Sphere, which is similar to the geometry we use.  More recent work
has explored the use of dual space~\cite{alignment2014,dualspace2013}
representations. Among the clustering-based approaches, Xu et
al.~\cite{kitware2013} improve this pipeline by introducing a new
point-line consistency function that models errors in the line segment
extraction step.

Alternatives to clustering-based approaches have been explored.
For example, vanishing point detection from line segments has been
modeled as an Uncapacitated Facility Location (UFL)
problem~\cite{global2013, selfsketch2012}. To avoid error accumulation
issues encountered by a step-by-step pipeline method, Barinova et
al.~\cite{geoparser2010} solve the problem in a unified framework,
where edges, lines, and vanishing points fit into a single graphical
model.

Our approach is motivated by the fact that properties of the scene,
including objects, can provide additional cues for vanishing point and
horizon line placement than line segments alone. Unlike existing
methods that use J-linkage~\cite{tardif2009, kitware2013} or similar
techniques to find an initial set of VPs by clustering detected lines
followed by a refinement step, our approach first proposes candidate
horizon lines using global image context.  


%
%
\subsection{Approach Overview}

Our approach is motivated by two observations: 1) traditional purely
geometric approaches to vanishing point detection often fail in
seemingly nonsensical ways and 2) identifying the true vanishing
points for many scenes is challenging and computationally expensive
due to the large number of outlier line segments. Driven by these
observations, we propose a two part strategy.  First, we use global
image context to estimate priors over the horizon line and the zenith
vanishing point (\secref{deeplearning}). Using these priors, we
introduce a novel VP detection method (\secref{precise}) that samples
horizon lines from the prior and performs a fast one-dimensional
search for high-quality vanishing points in each. Both steps are
essential for accurate results: the prior helps ensure a good
initialization such that our horizon-first detection method may obtain
very precise estimates that are necessary for many scene understanding
tasks. See \figref{pipeline} for an overview of our algorithm.

\section{Problem Formulation}
\label{sec:problem}

%

The goal of this work is to detect the horizon line, the zenith
vanishing point, and any horizontal vanishing points from a single
image. The remainder of this section defines the notation and basic
geometric facts that we will use throughout.  For clarity we use
unbolded letters for points in world coordinates or the image plane
and bolded letters for points or lines in homogeneous coordinates.  We
primarily follow the notation convention of Vedaldi and
Zisserman~\cite{selfsketch2012}.

Given a point $(u,v)$ in the image plane, its homogeneous coordinate
with respect to the calibrated image plane is denoted by:
\begin{displaymath}
  \myp = [\rho(u-c_u), \rho(v-c_v), 1]^\topt / \Sigma \; ,
\end{displaymath}
where $\rho$ is a scale constant, $(c_u, c_v)$ is the camera principal
point in the image frame, which we assume to be the center of the
image, and $\Sigma$ is the constant that makes $\myp$ a unit
vector.

\begin{figure}
  \centering
  \includegraphics[width=0.8\linewidth]{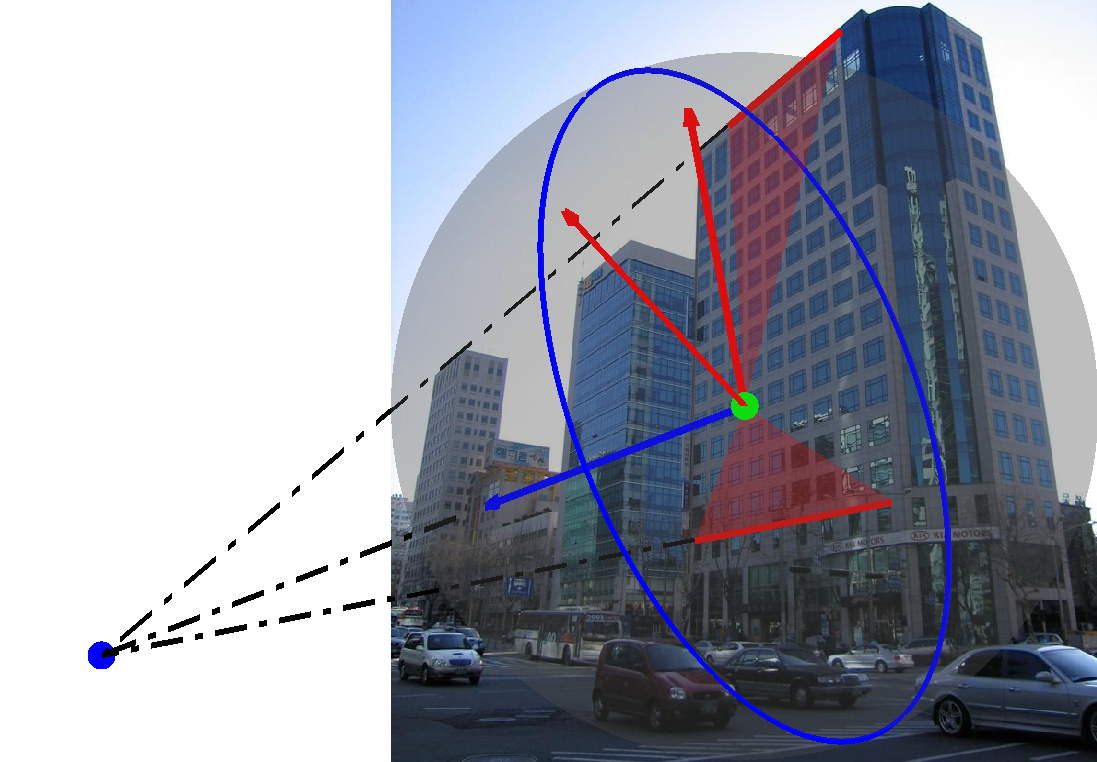}
  \caption{In homogeneous coordinates, lines (red lines) are defined
    by the normal (red arrow) of the plane (red triangle) they form
    with the origin (green dot). Two lines form a great circle (blue
    circle), whose normal (blue arrow) is their common point (blue
    dot) in homogeneous coordinates.}
  \label{fig:homoIntersection}
\end{figure}

In homogeneous coordinates, both lines and points are represented as
three-dimensional vectors (\figref{homoIntersection}). Computing the
line, $\myl$, that passes through two points, $(\myp_1, \myp_2)$, and
the point, $\myp$, at the intersection of two lines, $(\myl_1,
\myl_2)$, are defined as follows:
\begin{align}
  \myl = \frac{\myp_1 \times \myp_2}{\|\myp_1 \times
    \myp_2\|} 
  &&
  \myp = \frac{\myl_1 \times \myl_2}{\|\myl_1 \times \myl_2\|} \; .
\end{align}
We denote the smallest angle between two vectors $\mathbf{x}$ and
$\mathbf{y}$ with $\myang_{\mathbf{x},\mathbf{y}}
= |cos^{-1}(\mathbf{x}^\topt \mathbf{y})|$.
We use this to define the consistency between a line, $\myl$, and a
point,
$\myp$, as: $f_c(\myp, \myl) =
\operatorname*{max}(\theta_{con} - \myang_{\myp,\myl}, 0)$.
The maximum value of consistency between a vanishing point and a line
segment is $\theta_{con}$. This will occur if it is possible to extend
the line segment to contain the vanishing point.


\section{Horizon Priors from Global Image Context}
\label{sec:deeplearning}

Recent studies show that deep convolutional neural networks (CNNs) are
adaptable for a wide variety of tasks~\cite{yosinski2014transferable},
and are quite fast in practice.  We propose to use a CNN to extract
global image context from a single image.


We parameterize the horizon line by its slope angle, $\alpha \in
[-\pi, \pi)$, and offset, $o \in [0, \inf)$, which is the shortest
distance between the horizon line and the principal point. In order to
span the entire horizon line parameter space, we ``squash'' $o$ from
pixel coordinates to the interval $[0, \pi/2)$, through a one-to-one
function, $w = tan^{-1}(o / \kappa)$, in which $\kappa$ is a
scaling factor that affects how dense the sampling is near the center
of the image.



\begin{figure*}
  \hfill
  \includegraphics[width=0.19\linewidth]{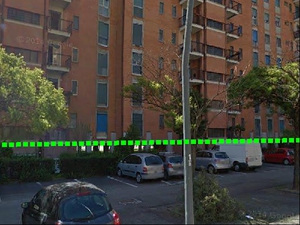}
  \hfill
  \includegraphics[width=0.19\linewidth]{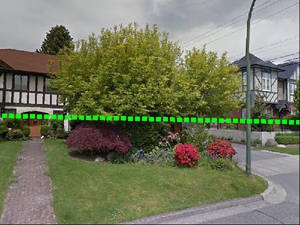}
  \hfill
  \includegraphics[width=0.19\linewidth]{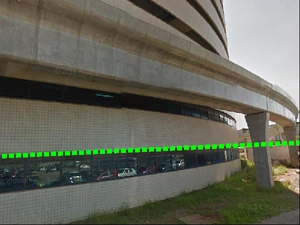}
  \hfill
  \includegraphics[width=0.19\linewidth]{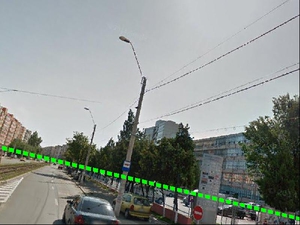}
  \hfill
  \includegraphics[width=0.19\linewidth]{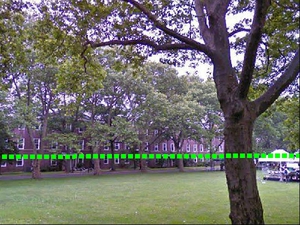}
  \hfill
  \caption{Example images from our training dataset
    (\secref{dataset}), each overlaid with the ground-truth horizon line.
  }
  \label{fig:dataset}
\end{figure*}

\subsection{Network Architecture}

For our task, we adapt the popular
AlexNet~\cite{krizhevsky2012imagenet} architecture, which was designed
for object recognition as part of the ImageNet ILSVRC-2012
challenge~\cite{russakovsky2015imagenet}. It consists of five
convolutional layers, each followed by a non-linearity (rectified
linear unit), and occasionally interspersed with pooling and local
response normalization. This is followed by three fully connected
layers (referred to as `fc6', `fc7', and `fc8'). A softmax is applied
to the final output layer to produce a categorical distribution over
1000 object classes.  We use this as a foundation to create a CNN that
simultaneously generates a categorical distribution for each
horizon-line parameter.

We modify the original AlexNet architecture in the following way: The
first five convolutional layers are left unmodified. These layers are
initialized with weights from a network trained for object detection
and scene classification~\cite{zhou2014places}. We remove the original
fully connected layers (`fc6'--`fc8') and add two disjoint sets of
fully connected layers (`fc6$\alpha$'--`fc8$\alpha$' and
`fc6$w$'--`fc8$w$'), one for each target label, $\alpha$ and $w$.  We
convert the slope, $\alpha$, and the squashed offset, $w$, into
independent categorical labels by uniformly dividing their respective
domains into 500 bins.  We randomly initialize the weights for these
new layers.

We train our network using stochastic gradient descent, with a
multinomial logistic loss function. The learning rates for the
convolutional layers are progressively increased such that the latter
layers change more. The new fully connected layers are given full
learning rate.

\subsection{Training Database}
\label{sec:dataset}
To support training our model of global image context, we construct a
large dataset of images with known horizon lines. We make use of
equirectangular panoramas downloaded from Google Street View in large
metropolitan cities around the world. We identified a set of cities
based on population and Street View coverage. From each city, we
downloaded panoramas randomly sampled in a $5km \times 5km$ region
around the city center. This resulted in $11\,001$ panoramas from 93
cities. Example cities include New York, Rio de Janeiro, London, and
Melbourne.

We extracted 10 perspective images from each panorama with randomly
sampled horizontal field-of-view (FOV), yaw, pitch, and roll. Here yaw
is relative to the Google Street View capture vehicle. We sampled
horizontal FOV from a normal distribution with $\mu=60^\circ$ and
$\sigma=10^\circ$. Similarly, pitch and roll are sampled from normal
distributions with $\mu=0^\circ$ and $\sigma=10^\circ$ and
$\sigma=5^\circ$, respectively. Yaw is sampled uniformly.  We truncate
these distributions such that horizontal FOV $\in [40^\circ,
80^\circ]$, pitch $\in [-30^\circ, 30^\circ]$, and roll $\in
[-20^\circ, 20^\circ]$. These settings were selected empirically to
match the distribution of images captured by casual photographers in
the wild.

Given the FOV, pitch, and roll of a generated perspective image, it is
straightforward to compute the horizon line position in image space.
In total, our training database contains $110\,010$ images with known
horizon line.  \figref{dataset} shows several example images from
our dataset annotated with the ground-truth horizon line.

\subsection{Making the Output Continuous}

\begin{figure}
 \centering
 \includegraphics[width=.48\linewidth]{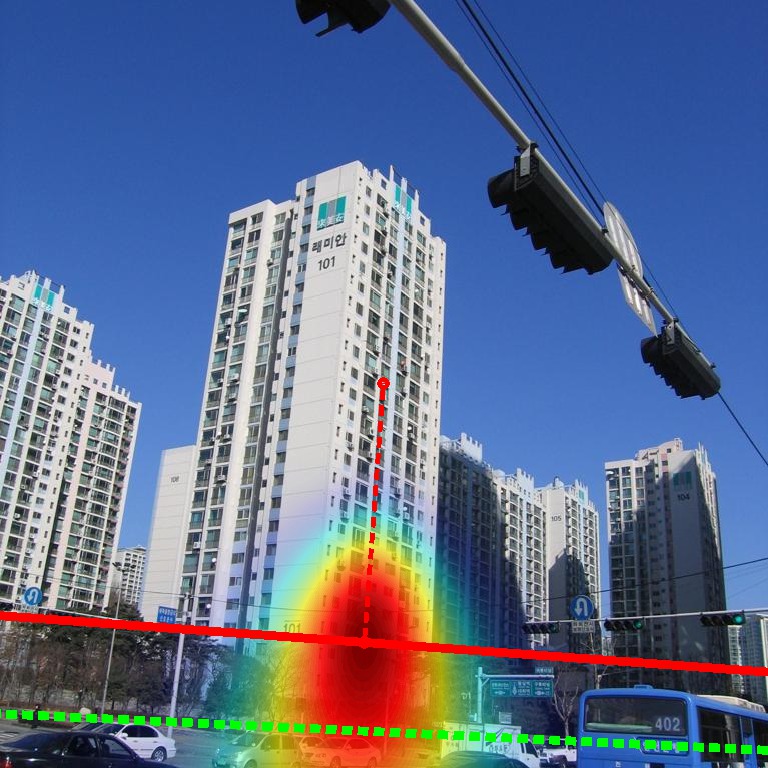}
 \includegraphics[width=.48\linewidth]{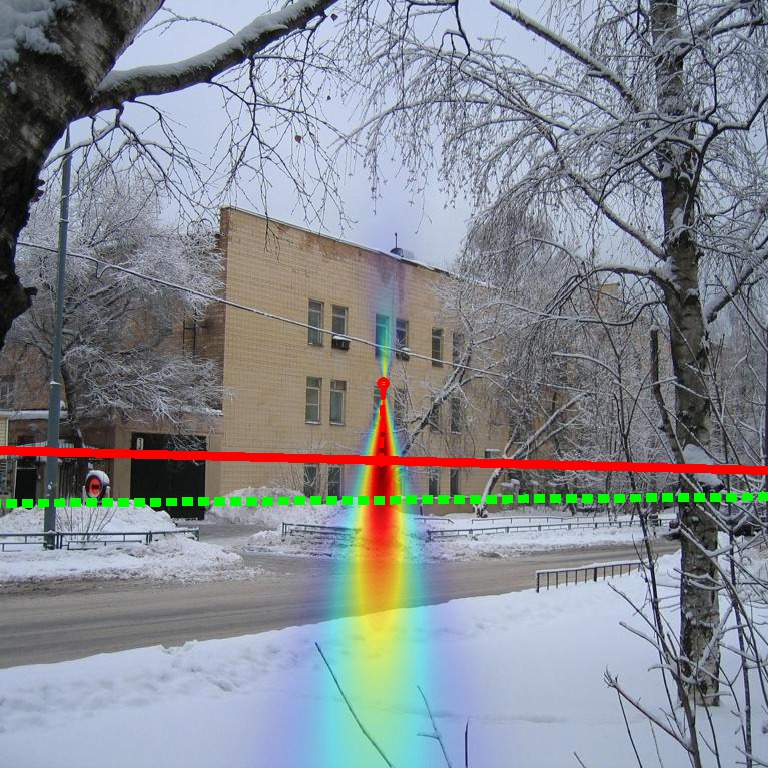}

 \caption{Global image context imposes a strong prior on horizon line
   location. The output of our CNN is visualized as an overlaid
   heatmap, with red indicating more likely locations. For each image,
   the ground-truth horizon line (dash green) and the line that
   maximizes the prior (red) are shown.}

 \label{fig:horpdf}
\end{figure}

Given an image, $I$, the network outputs a categorical probability
distribution for the slope, $\alpha$, and squashed offset, $w$. We
make these distributions continuous by approximating them with a
Gaussian distribution. For each, we estimate the mean and variance
from $5\,000$ samples generated from the categorical probability
distribution. Since the relationship between $w$ and $o$ is
one-to-one, this also results in a continuous distribution over
$o$. The resulting distributions, $p(\alpha \given I)$ and $p(o \given
I)$, are used in the next step of our approach to aid in detecting the
zenith VP and as a prior for sampling candidate horizon
lines. To visualize this distribution we observe that the horizon line
can be uniquely defined by the point on the line closest to the
principal point. Therefore, we can visualize a horizon line
distribution as a distribution over points in the image.
\figref{horpdf} shows this distribution for two images.

\section{Horizon-First Vanishing Point Detection}
\label{sec:precise}

We propose an approach to obtain accurate estimates of the horizon
line, the zenith vanishing point, and one or more horizontal vanishing
points. Given an image, our approach makes use of the distributions
estimated from global image context (\secref{deeplearning}) and
line segments extracted with LSD~\cite{lsd2010}. The algorithm
consists of the following major steps:
\begin{compactenum} 

  \item detect the zenith vanishing point
    (\secref{zenithdetection})

  \item \label{alg:findhorvp} detect horizontal vanishing points on
    horizon line candidates
    (\secref{horizondetection})

  \item \label{alg:scorehorizon} score horizon line candidates with
    horizontal vanishing points (\secref{horizonscoring})

\end{compactenum}
The remainder of this section provides details for each of these steps.

\subsection{Detecting the Zenith Vanishing Point}
\label{sec:zenithdetection}


To detect the zenith vanishing point, we first select an initial set
of line segments using the zenith direction,
$\myl_\myz$, from the global image context, then use
the RANSAC~\cite{ransac1981} algorithm to refine it.  The zenith direction
is the line connecting the principal point and the zenith vanishing
point, which is uniquely determined by the horizon line slope (see
supplemental material for a proof).

We compute our initial estimate of $\myl_{\myz}$ using the global
image context by choosing the value that maximizes the posterior:
$\hat{\alpha} = \operatorname*{arg\,max}_\alpha p(\alpha \given I)$.
To handle the presence of outlier line segments, we first select a set
of candidate vertical line segments as the RANSAC inputs by
thresholding the angle between each line segment and the estimated
zenith direction, $\myang_{\myl,\myl_{\myz}} < \theta_{ver}$.  For a
randomly sampled pair of line segments with intersection, $\myp$, we
compute the set of inlier line segments, $\{\myl \mid f_c(\myp, \myl)
> 0\}$.  If the largest set of inliers has a sufficient portion (more
than 2\% of candidate line segments), we obtain the final estimate of
the zenith vanishing point, $\mathbf{z}$, by minimizing the algebraic
distance, $\|\myl^\topt \myp\|$ using singular value decomposition
(SVD), and update the zenith direction, $\myl_{\myz}$. Otherwise, we
keep the zenith direction estimated from the global image context.

%
%

\subsection{Detecting Horizontal Vanishing Points}
\label{sec:horizondetection}

\begin{figure}
  \centering
  \includegraphics[width=.4075\linewidth]{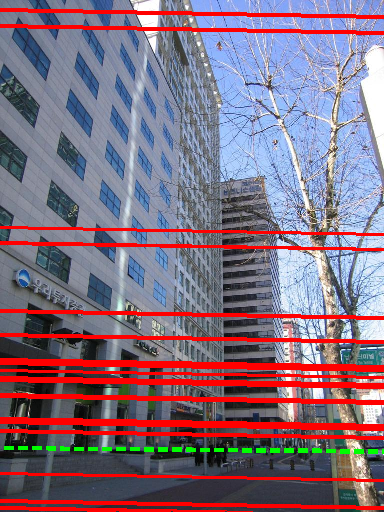}
  \includegraphics[width=.4075\linewidth]{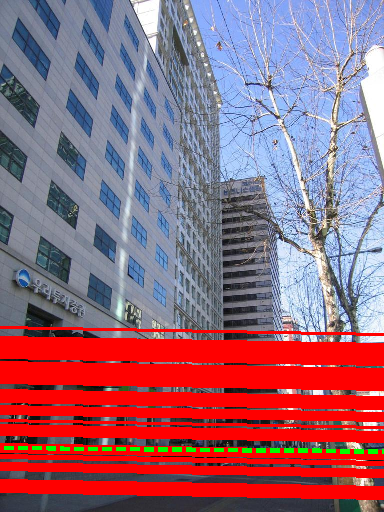}
  \includegraphics[width=.1630\linewidth]{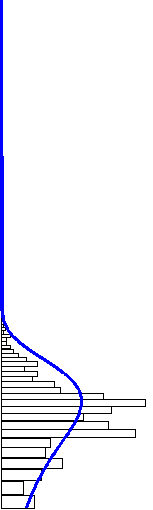}
  \caption{Our method samples more horizon line candidates (red) near
    the ground truth (green dash) with (middle) global image context
    than without (left).  In the case of sampling with global image
    context, the offset PDF, $p(o|I)$ (blue curve), is fit from the
    CNN categorical probability distribution outputs (hollow
    bins). For clarity, we only show a reduced number of horizon line
    candidates and bins.}
  \label{fig:horsampling}
\end{figure}

We start with sampling a set of horizon line candidates,
$\{\myh_i\}_1^S$, that are perpendicular to $\myl_{\myz}$ in the image
space, under the distribution of horizon line offsets, $p(o | I)$.
See \figref{horsampling} for examples of horizon line sampling with
and without global context.

For each horizon line candidate, we identify a set of horizontal VPs
by selecting points along the horizon line where many line segments
intersect. We assume that for the true horizon line the identified
horizontal VPs will be close to many intersection points and that
these intersections will be more tightly clustered than for
non-horizon lines. We use this intuition to define a scoring function
for horizon line candidates.

As a preprocessing step, given the zenith direction, $\myl_{\myz}$,
and a horizon line candidate, $\myh$, we filter out nearly vertical
line segments ($\myang_{\myl,\myl_{\myz}} < \theta_{ver}$), which are
likely associated with the zenith vanishing point, and nearly
horizontal line segments ($\myang_{\myl,\myh} < \theta_{hor}$), which
result in noisy horizon line intersection points.  We remove such
lines from consideration because they lead to spurious, or
uninformative, vanishing points, which decreases accuracy.

Given a horizon line candidate, $\myh$, and the filtered line segments
in homogeneous coordinates, $\mathcal{L} = \{\myl_i\}$, we select
a set of horizontal VPs, $\mathcal{P} = \{\myp_i\}$, by minimizing
the following objective function:
\begin{align} 
  g(\mathcal{P} \given \myh, \mathcal{L}) & = -\sum_{\myp_i
    \in \mathcal{P}} {\sum_{\myl_j \in
      \mathcal{L}}f_c(\myp_i, \myl_j)}
  \label{eq:obj} \\
  \text{subject to:} \nonumber \\
   & \myang_{\myp_i,\myp_j} > \theta_{dist}
  \; \text{and} \; \left< \mathbf{p}_i,\mathbf{h} \right>=0, 
  \; \forall (i,j) \nonumber \; .
\end{align} 
%
The constraint prevents two vanishing points from being too close
together, which eliminates the possibility of selecting multiple
vanishing points in the same location.

We propose the following combinatorial optimization process for
obtaining an initial set of vanishing points, followed by a
constrained nonlinear optimization to refine the vanishing points.

\subsubsection{Initialization by Random Sampling and Discrete
Optimization}
\label{sec:initialization}

To choose an initial set of candidate vanishing points,
$\{\myp_i\}_1^M$, we randomly select a subset of line segments,
$\{\myl_i\}_1^M$, and compute their intersection with the horizon
line.  We then construct a graph with a node for each vanishing point,
$\myp_i$, each with weight $\sum_{\myl_j \in \mathcal{L}}f_c(\myp_i,
\myl_j)$, which is larger if there are many line segments in the image
that are consistent with $\myp_i$. Pairs of nodes, $(i,j)$, are
connected if the corresponding vanishing points, $\myp_i, \myp_j$, are
sufficiently close in homogeneous space ($\myang_{\myp_i,\myp_j} \le
\theta_{dist}$).

\begin{figure}
  \centering
  \includegraphics[width=.92\linewidth]{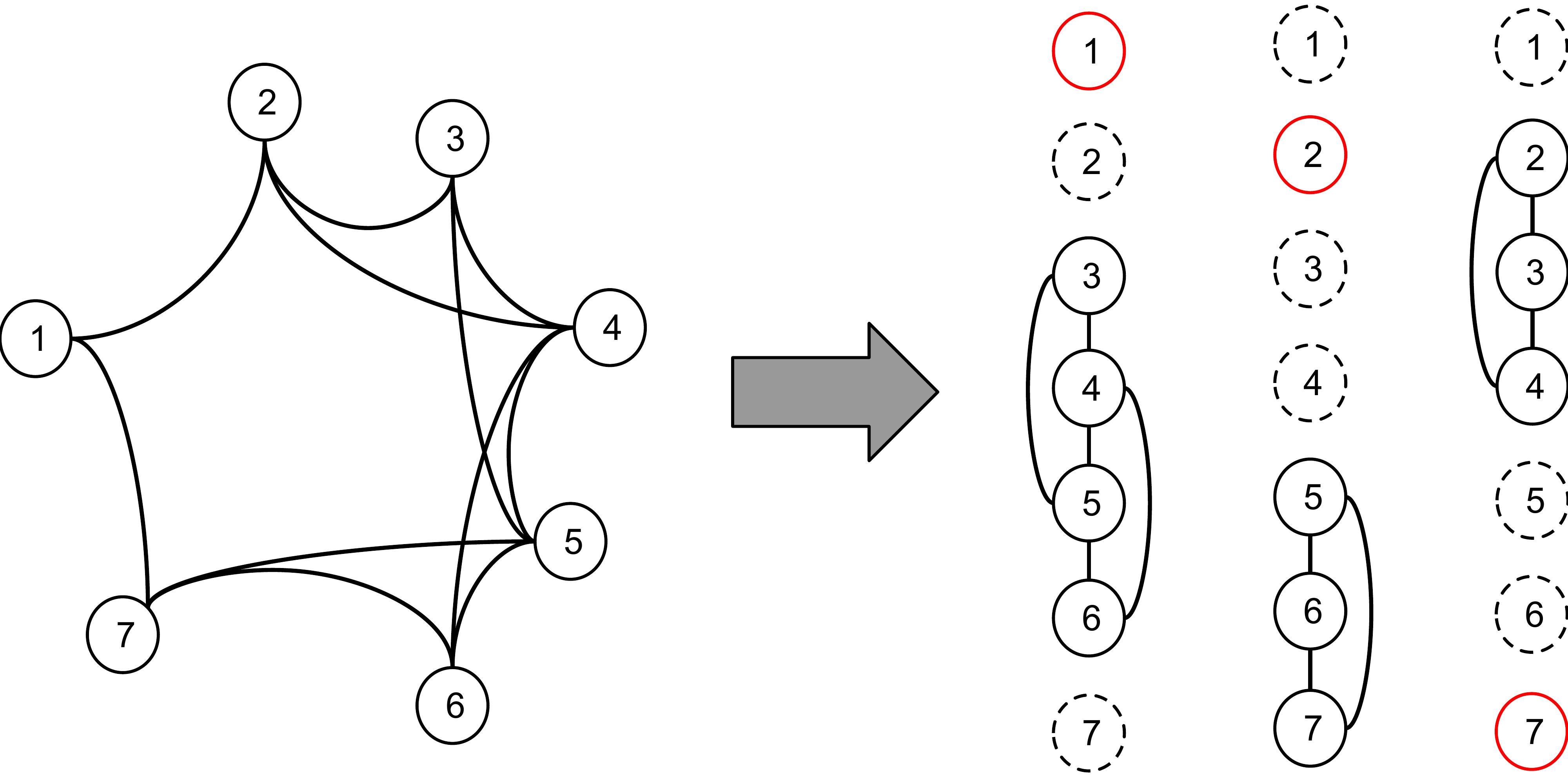}

  \caption{A ring-like graph (left) is converted into three nearly
    linear subgraphs (right) by partitioning around a node with
    minimal degree. For the subgraphs, the red node is mandatory, the
    dashed nodes are excluded, and a subset of the solid nodes are
    selected using dynamic programming.}

  \label{fig:subset}
\end{figure}

From this randomly sampled set, we select an optimal subset of VPs by
maximizing the sum of weights, while ensuring no VPs in the final set
are too close.  Therefore, the problem of choosing the initial set of
VPs reduces to a maximum weighted independent set problem, which is
NP-hard in general. Due to the nature of the constraints, the
resulting graph has a ring-like structure which means that, in
practice, the problem can be quickly solved. Our solver exploits this
sparse ring-like structure by finding a set of VPs that when removed
convert the ring-like graph into a set of nearly linear sub-graphs
(\figref{subset}). We solve each subproblem using dynamic
programming. The set of VPs with maximum weight, $\{\myp_i\}_{opt}$,
is used as initialization for local refinement. Usually, 2--4 such
vanishing points are found near the horizon line ground truth.

\subsubsection{Vanishing Points Refinement}
\label{sec:refinement}

Since they were randomly sampled, the set of vanishing points
selected during initialization, $\{\myp_i\}_{opt}$, may not be at the
optimal locations.  We optimize their locations to further minimize
the objective function \equref{obj}. We perform an EM-like algorithm
to refine the vanishing point locations, subject to the constraint
that they lie on the horizon line:
\begin{compactitem}

\item \emph{E-step}: Given a vanishing point, $\myp$, assign
  line segments that have positive consistency with $\myp$:
  $\{\myl|f_c(\myp,\myl) > 0\}$.
\item \emph{M-step}: Given the assigned line segments as a matrix,
  $\myL = [\myl_1, \myl_2,\dots ,\myl_n]$, and the horizon line,
  $\myh$, both represented in homogeneous coordinates, we 
  solve for a refined vanishing point, $\myp^*$, by minimizing the algebraic distance, $\|
  \myL^\topt \myp \|$ such that $\myh^\topt \myp = 0$. We define a
  basis, $\myB_\myh$, for the null space of $\myh$, and reformulate
  the problem as $\mathbf{\lambda}^* = \operatorname*{arg\,min} \|
  \myL^\top \myB_\myh \mathbf{\lambda} \|$, which we solve using SVD.
  Given the optimal coefficients, $\mathbf{\lambda}^*$, we reconstruct
  the optimal vanishing point as: $\myp^* =
  \frac{\myB_\myh\mathbf{\lambda}^*}
  {\|\myB_\myh\mathbf{\lambda}^*\|}$.

\end{compactitem}

We run this refinement iteration until convergence. In practice, this
converges quickly; we run at most three iterations for all the
experiments. The final set of optimized VPs is then used to assign a
score to the current horizon line candidate.

\subsection{Optimal Horizon Line Selection}
\label{sec:horizonscoring}

For each horizon line candidate, we assign a score based on the total
consistency of lines in the image with the VPs selected in the
previous section.  The score of a horizon line candidate, $\myh$, is
defined as:
\begin{equation}
  score(\myh) = \sum_{\{\tilde{\myp}_i\}} \sum_{\myl_j \in
    \mathcal{L}}f_c(\tilde{\myp}_i, \myl_j) \; .
\end{equation}
To reduce the impact of false positive vanishing points, we select
from $\{\myp_i\}_{opt}$ the two highest weighted vanishing points (or
one if $\{\myp_i\}_{opt}$ contains only one element),
$\{\tilde{\myp}_i\}$, for horizon line scoring.

%
%
\section{Evaluation}
\label{sec:evaluation}

We perform an extensive evaluation of our methods, both quantitatively
and qualitatively, on three benchmark datasets. The results show that
our method achieves state-of-the-art performance based on horizon-line
detection error, the standard criteria in recent work on VP
detection~\cite{geoparser2010,alignment2014,selfsketch2012,kitware2013}.
Horizon detection error is defined as the maximum distance from the
detected horizon line to the ground-truth horizon line, normalized by
the image height.  Following tradition, we show the cumulative
histogram of these errors and report the area under the curve (AUC).

Our method is implemented using MATLAB, with the exception of
detecting line segments, which uses an existing C++
library~\cite{lsd2010}, and extracting global image context, which we
implemented using Caffe~\cite{jia2014caffe}. We use the parameters
defined in \tabref{parameters} for all experiments. This differs from
other methods which usually use different parameters for different
datasets.

\begin{table}[h]
  \centering
  \caption{
   Algorithm parameters (given an $H\times W$ image).
  }
  \vspace{-.5em}
  \begin{tabular}{|c|l|l|}
    \hline
    Name & Usage(s) & Value\\
    \hline
    $\theta_{con}$ & \secref{problem} & $2^\circ$\\
    $\rho$ & \secref{problem} & $2/\operatorname{max}(H,W)$ \\
    $\kappa$ & \secref{deeplearning} & $ 1/5 \times H$ \\
    $\theta_{ver}$ & \secref{zenithdetection}, \secref{horizondetection} & $ \myang_{\myl,\myl_{\myz}} < 10^\circ$ \\
    $\theta_{hor}$ & \secref{horizondetection} & $\myang_{\myl,\myh} <
    1.5^\circ$ \\
    $S$ & \secref{horizondetection} & 300 candidates \\
    $M$ & \secref{initialization} & 20 line segments \\
    $\theta_{dist}$ & \secref{horizondetection}, \secref{initialization} & $
    \myang_{\myp_i,\myp_j} > 33^\circ$\\
    \hline
  \end{tabular}
  \label{tab:parameters}
\end{table}

\begin{figure*}[]
  \centering
  \begin{subfigure}{.30\linewidth}
    \centering
    \includegraphics[width=\linewidth]{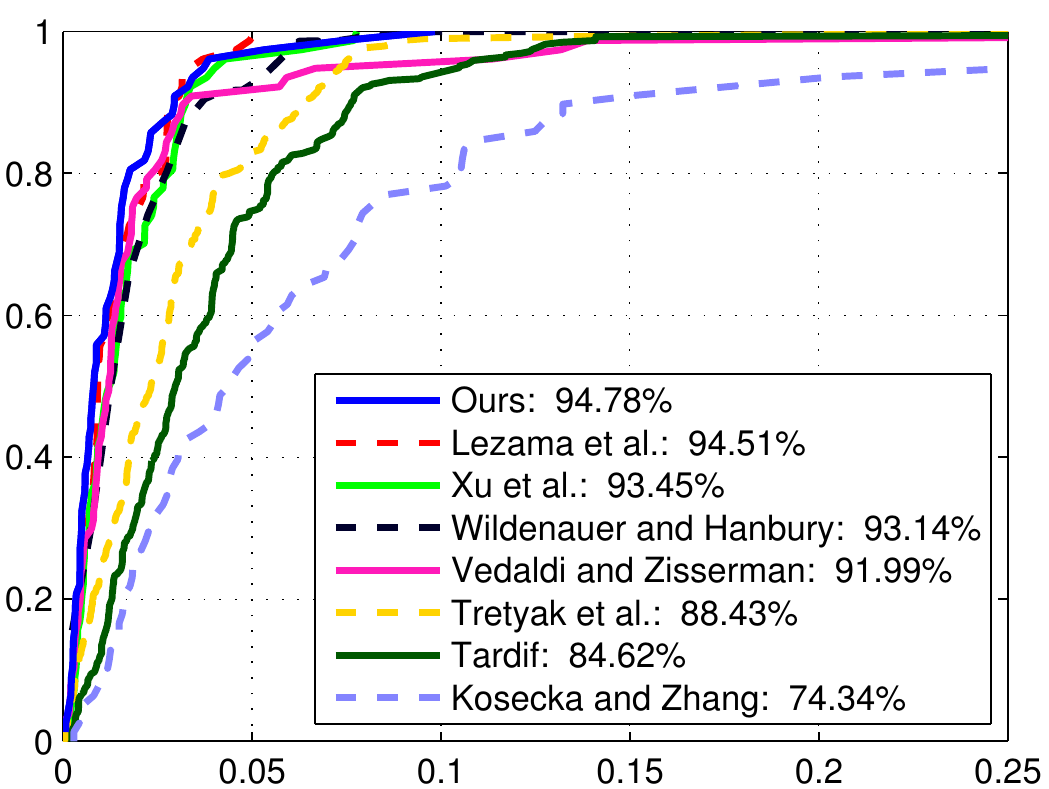}
    \subcaption{YUD}
    \label{fig:horError_YUD}
  \end{subfigure}
  \hfill
  \begin{subfigure}{.3\linewidth}
    \centering
    \includegraphics[width=\linewidth]{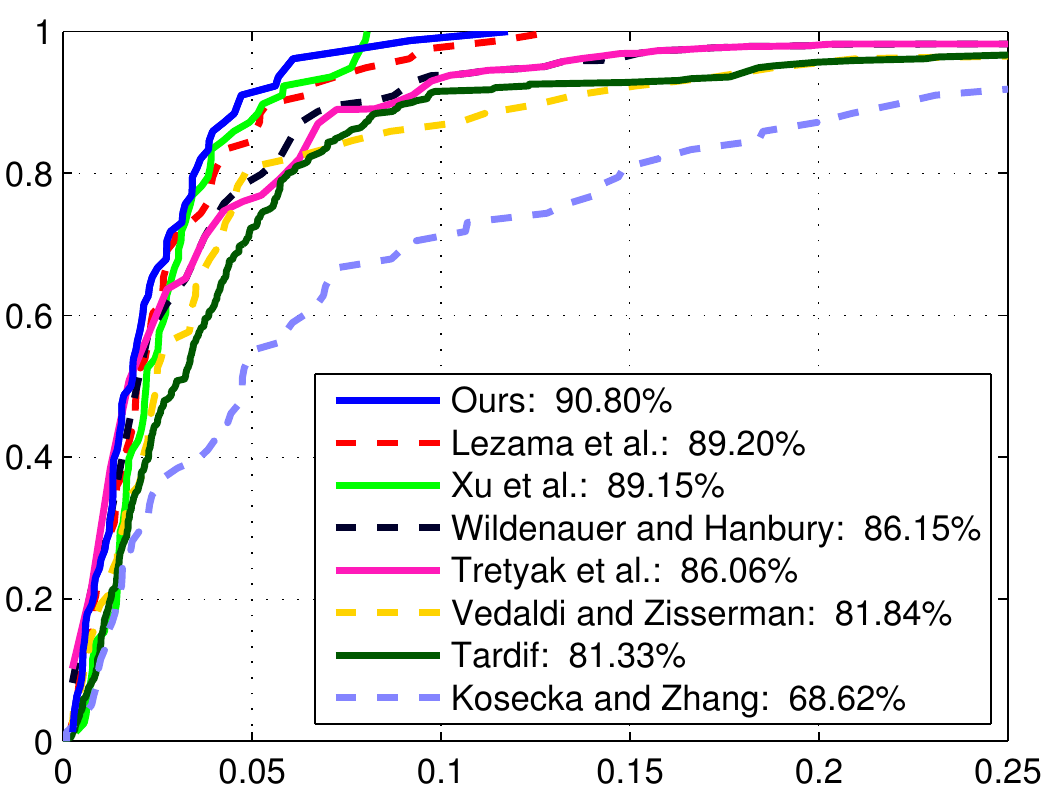}
    \subcaption{ECD}
    \label{fig:horError_ECD}
  \end{subfigure}
  \hfill
  \begin{subfigure}{.3\linewidth}
    \centering
    \includegraphics[width=\linewidth]{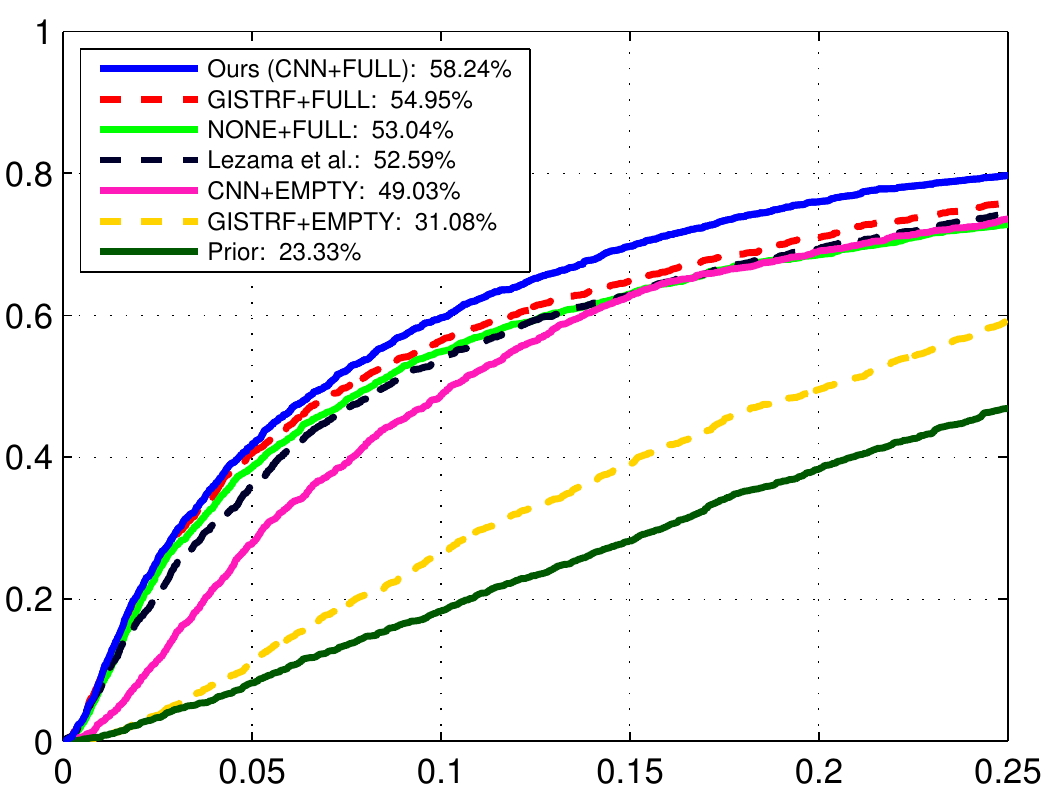}
    \subcaption{HLW}
    \label{fig:hlw_err}
  \end{subfigure}
  \caption{For three benchmark datasets, the fraction of images
    (y-axis) with a horizon error less than a threshold (x-axis). The
    AUC for each curve is shown in the legend. For additional details see \secref{evaluation}.}
  \label{fig:horError}
\end{figure*}

\subsection{Quantitative Evaluation}


The York Urban Dataset (YUD)~\cite{edgebased2008} is a commonly used dataset
for evaluating horizon line estimation methods. It contains 102 images
and ground-truth vanishing points. The scenes obey the Manhattan-world
assumption, however we do not take advantage of this assumption.
\figref{horError_YUD} shows the performance of our methods relative to
previous work on YUD. These results demonstrate that our method
achieves state-of-the-art AUC, improving upon the previous best of
Lezama et al.~\cite{alignment2014} by 0.28\%, a relative
improvement\footnote{We define the relative improvement as
$\frac{\text{AUC}_{new} - \text{AUC}_{old}}{1 - \text{AUC}_{old}}$.} of
5\%. This is especially impressive given that our method only requires
an average of 1 second per image, while Lezama et al.\ requires
approximately 30 seconds per image.


The Eurasian Cities Dataset (ECD)~\cite{geoparser2010} is another
commonly used benchmark dataset, which is considered challenging due
to the large number of outlier line segments and complex scene
geometries. It contains 103 images captured in urban areas and, unlike
the YUD dataset, not all images satisfy the Manhattan-world
assumption. It provides reliable horizon line ground truth and is
widely considered difficult for horizon line detection. To our
knowledge, the previous state-of-the-art performance in terms of the
AUC metric on this dataset was achieved by Lezama et
al.~\cite{alignment2014}. Our algorithm improves upon their
performance, increasing the state of the art to 90.8\%.  This is a
significant relative improvement of 14.8\%, especially considering
their improvement relative to the state of the art was 0.5\%.  On ECD,
our method takes an average of 3 seconds per image, while Lezama et
al.\ requires approximately 60 seconds per image. We present the
performance comparison with other methods in \figref{horError_ECD}.


The Horizon Lines in the Wild (HLW)
dataset~\cite{authors2016deephorizon} is a new, very challenging
benchmark dataset. We use the provided test set, which contains
approximately $2\,000$ images from diverse locations, with many images
not adhering to the Manhattan-world assumption. \figref{hlw_err}
compares our method with the method of Lezama et
al.~\cite{alignment2014} (the only publicly available implementation
from a recent method). Our method is significantly better, achieving
58.24\% versus 52.59\% AUC.

\begin{table}[t] \centering
  \caption{Component error analysis (AUC).}
  \vspace{-.5em}
  \begin{tabular}{|l|c|c|c|}
    \hline
    Method & YUD & ECD & HLW \\
    \hline
    \hline
    Lezama et al.~\cite{alignment2014} & 94.51\% & 89.20\% & 52.59\% \\
    \hline
    NONE+FULL & 93.87\% & 87.94\% & 53.04\% \\
    \hline
    GISTRF+EMPTY & 53.36\% &  32.69\% & 31.08\% \\
    GISTRF+FULL & 94.66\% &  87.60\% & 54.95\% \\
    \hline
    CNN+EMPTY & 73.67\% &  67.64\% & 49.03\% \\
    CNN+FULL (Ours) & {\bf 94.78\%} & {\bf 90.80\%} & {\bf 58.24\%} \\
    \hline
  \end{tabular}
  \label{tab:contribution}
\end{table}

\subsection{Component Error Analysis}
\label{sec:analysis}

Our method consists of two major components: global context extraction
(\secref{deeplearning}) and horizon-first vanishing point detection
(\secref{precise}). This section provides an analysis of the impact
each component has on accuracy. 

To evaluate the impact of global context extraction, we considered
three alternatives: our proposed approach (CNN), replacing the CNN
with a random forest (using the Python ``sklearn'' library with 25 trees) applied to a GIST~\cite{oliva2001modeling}
descriptor (GISTRF), and omitting context entirely
(NONE). When omitting the global context, we assume no camera
roll (horizon lines are horizontal in the image) and sample horizon
lines uniformly between $[-2H,2H]$ ($H$ is the image height).  To
evaluate the impact of vanishing point detection, we considered two
alternatives: our proposed approach (FULL) and omitting the vanishing
point detection step (EMPTY). When omitting vanishing point detection,
we directly estimate the horizon line, $(\alpha, o)$, by maximizing
the posterior estimated by our global-context CNN, $p(\alpha, o | I)$.

\begin{figure}
  \centering
  \includegraphics[width=.48\linewidth]{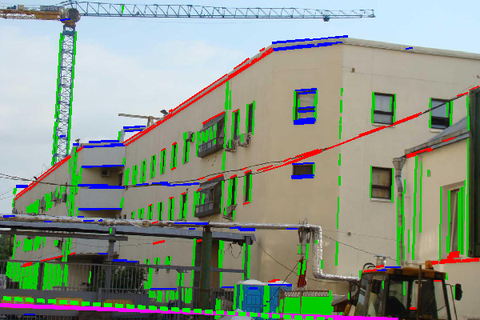}
  \includegraphics[width=.48\linewidth]{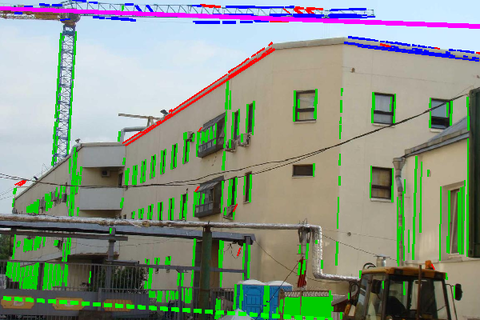}
  \includegraphics[width=.48\linewidth]{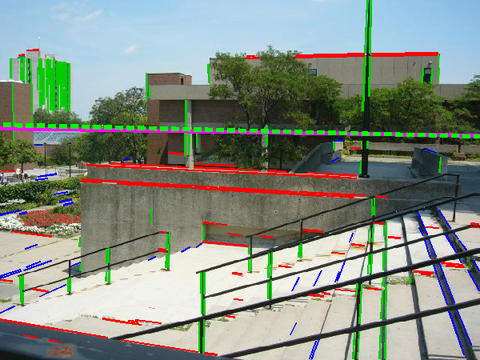}
  \includegraphics[width=.48\linewidth]{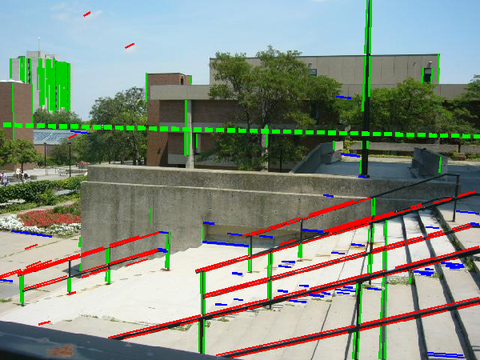}
  \caption{Two images where horizon line estimates are much better with global context (left) than without (right).}
  \label{fig:iscontext}
\end{figure}

Quantitative results presented in \tabref{contribution} show that both
components play important roles in the algorithm and that CNN provides
better global context information than GISTRF.  Though our vanishing
point detection performs well by itself (see column NONE+FULL), global
image context helps improve the accuracy further. \figref{hlw_err}
visualizes these results as a cumulative histogram of horizon error
on HLW.  To illustrate the impact of global image context, we present
two examples in \figref{iscontext} that compare horizon line
estimates obtained using global context (CNN+FULL) and without
(NONE+FULL).  When using global context, the estimated horizon lines
are very close to the ground truth.  Without, the estimates obtained
are implausible, even resulting in an estimate that is off the image.


\begin{figure*}
  \centering
  \includegraphics[width=0.1571\linewidth]{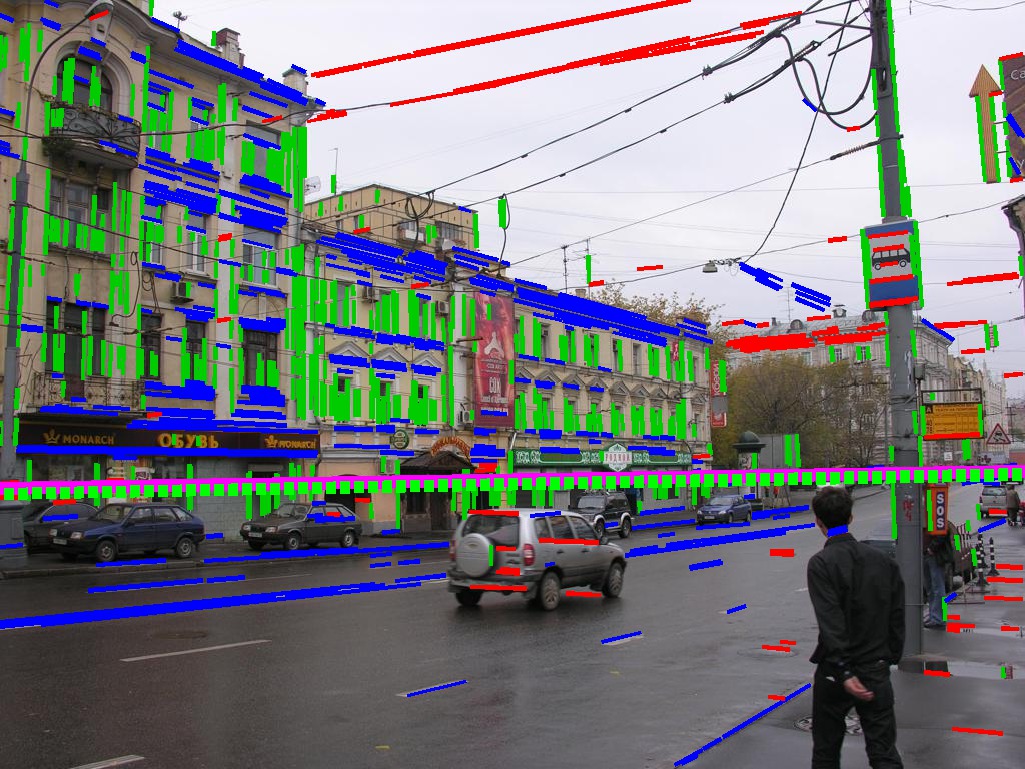}
  \includegraphics[width=0.0884\linewidth]{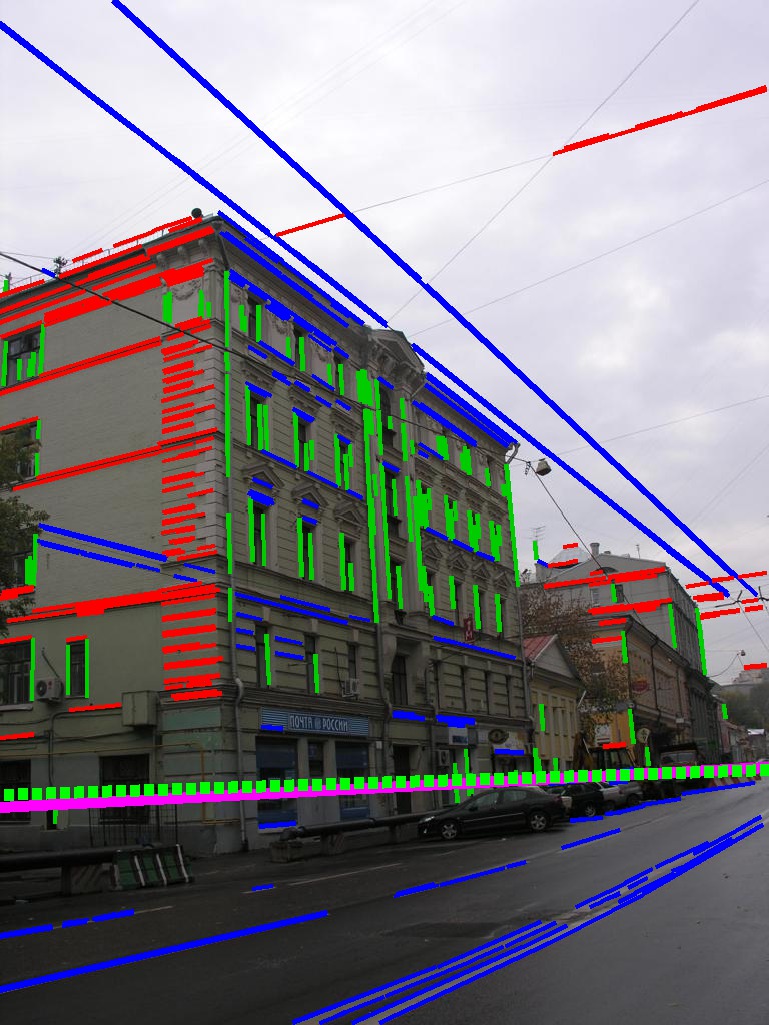}
  \includegraphics[width=0.1572\linewidth]{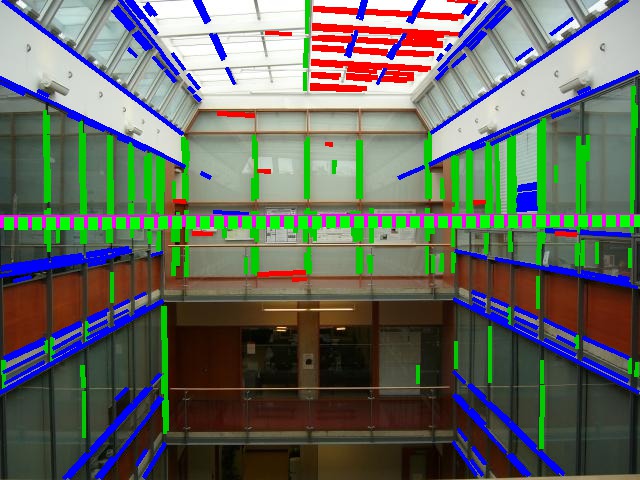}
  \includegraphics[width=0.1571\linewidth]{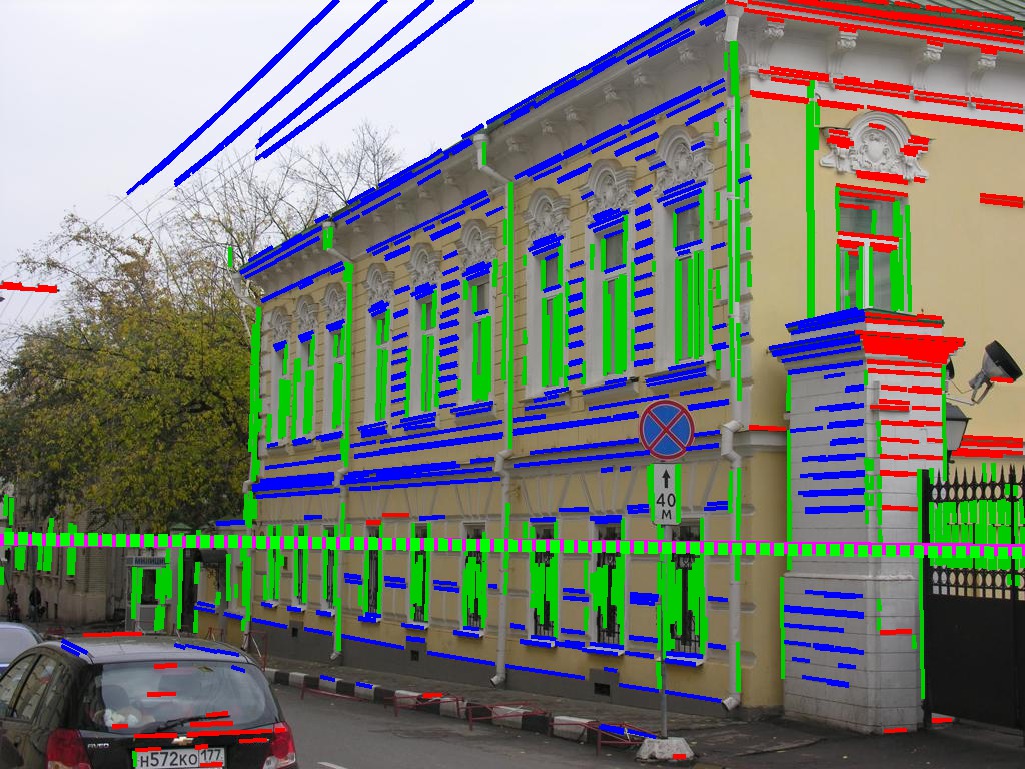}
  \includegraphics[width=0.1572\linewidth]{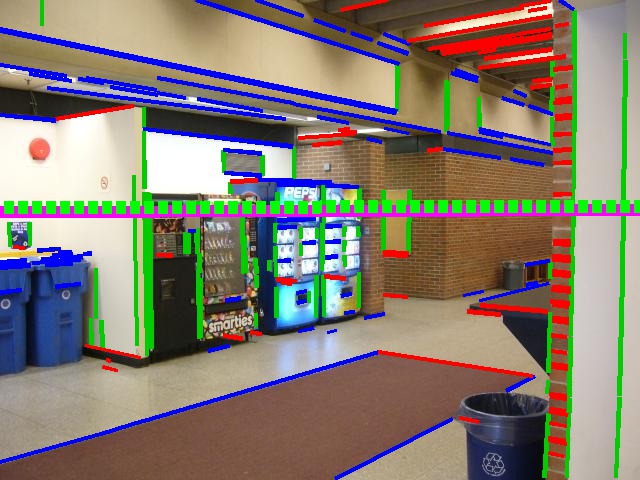}
  \includegraphics[width=0.0884\linewidth]{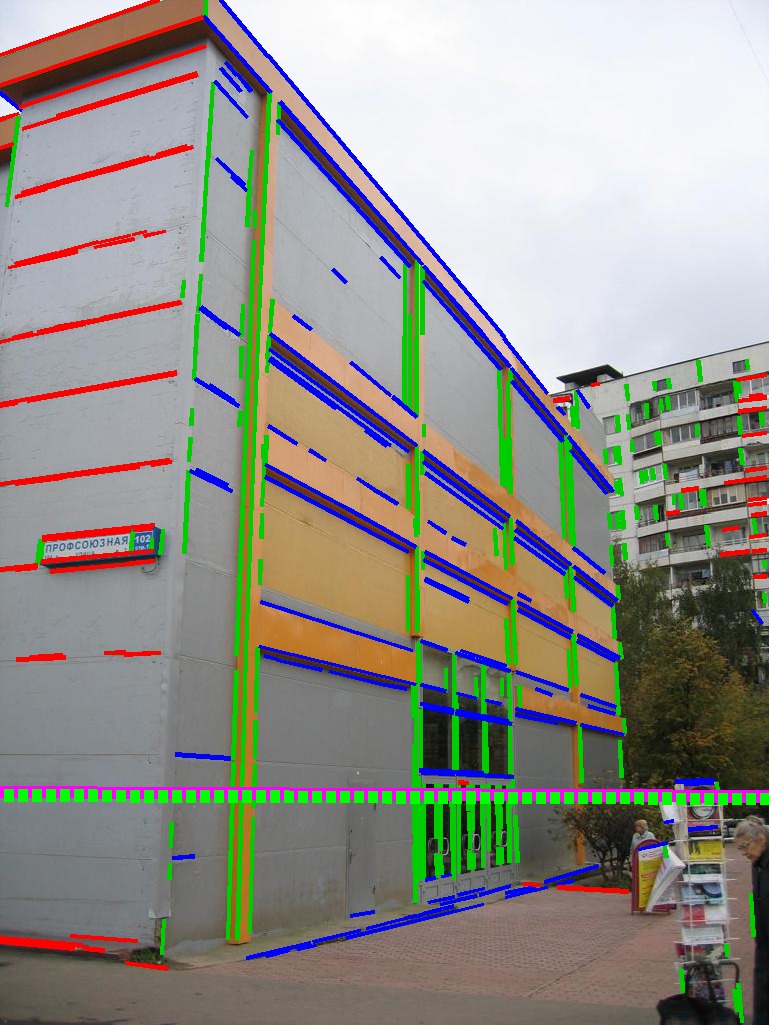}
  \includegraphics[width=0.1571\linewidth]{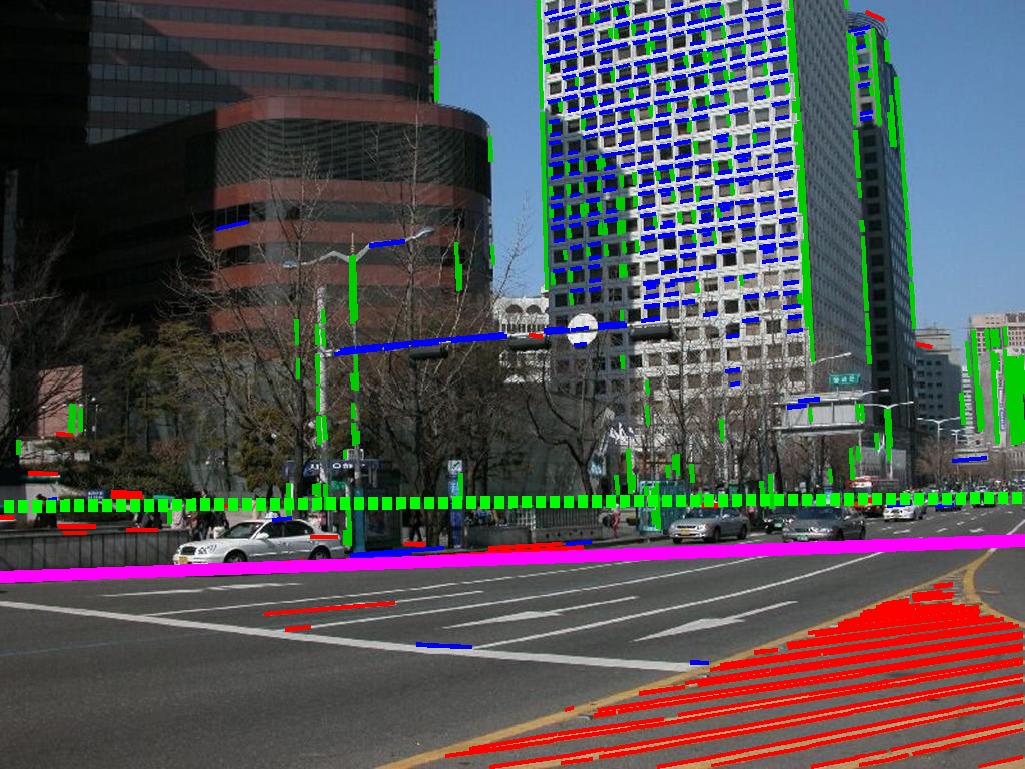}
  \includegraphics[width=0.1375\linewidth]{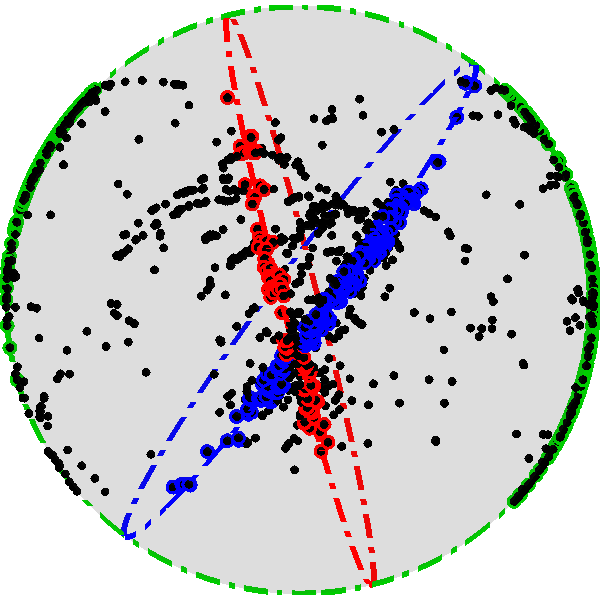}
  \includegraphics[width=0.1375\linewidth]{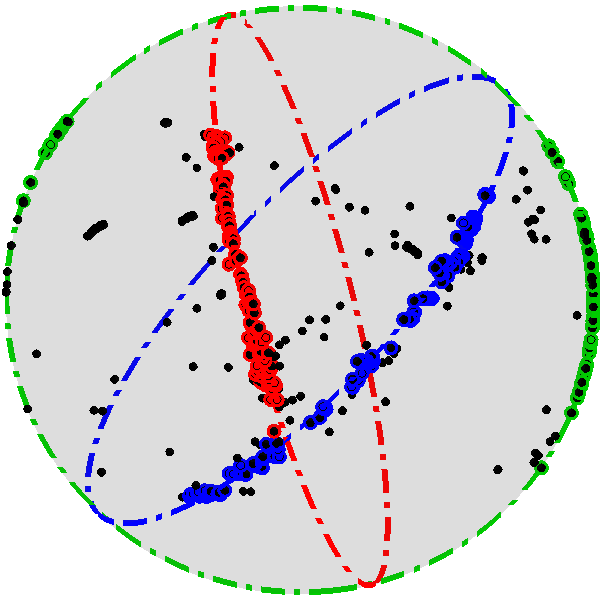}
  \includegraphics[width=0.1375\linewidth]{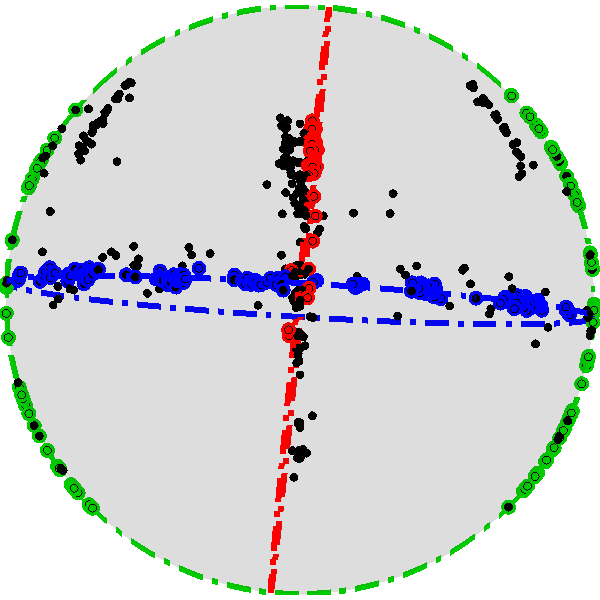}
  \includegraphics[width=0.1375\linewidth]{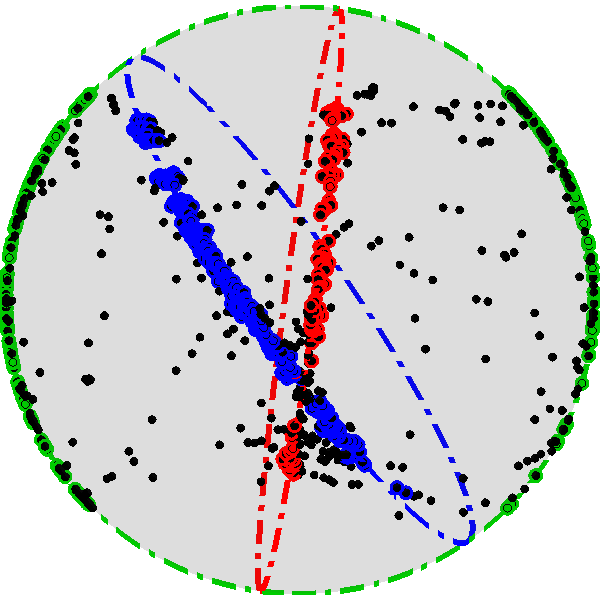}
  \includegraphics[width=0.1375\linewidth]{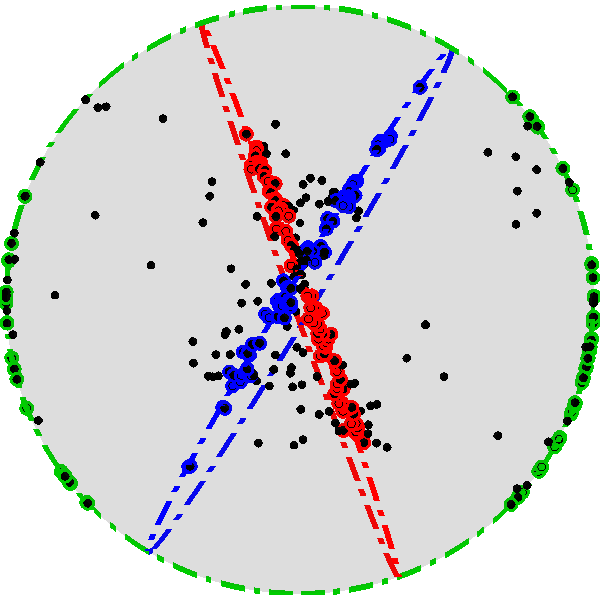}
  \includegraphics[width=0.1375\linewidth]{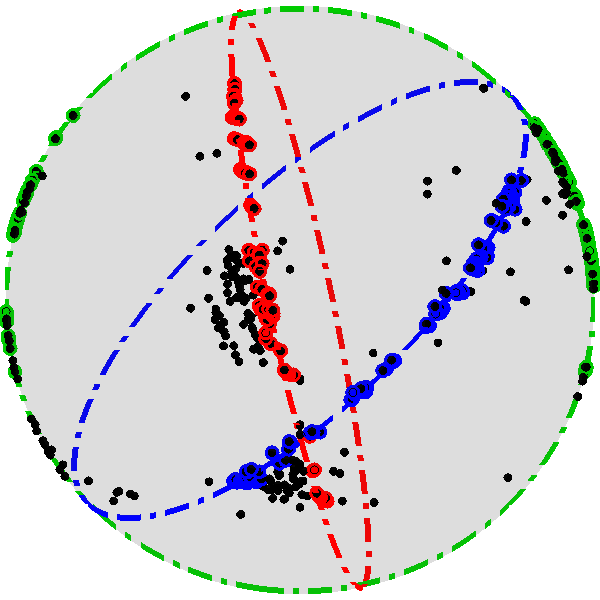}
  \includegraphics[width=0.1375\linewidth]{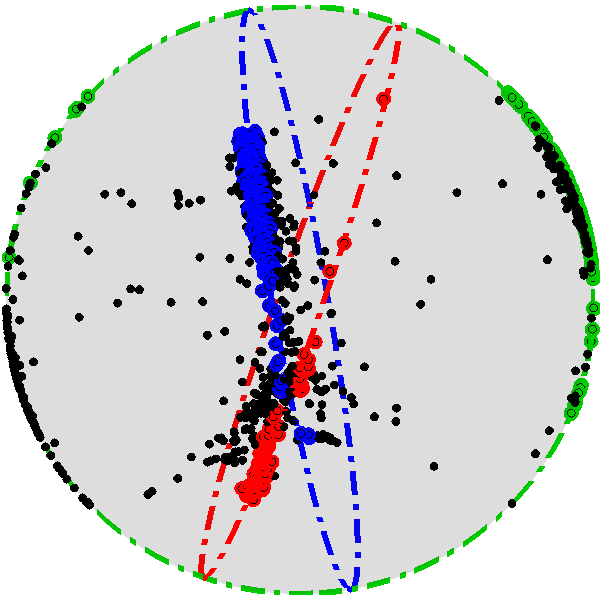}
  
  \includegraphics[width=0.1540\linewidth]{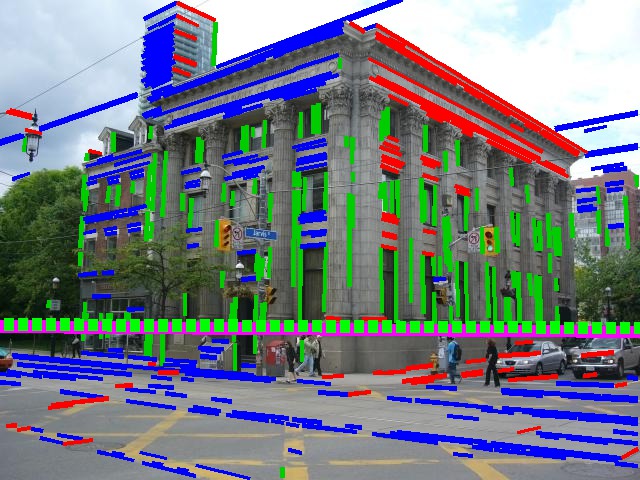}
  \includegraphics[width=0.0867\linewidth]{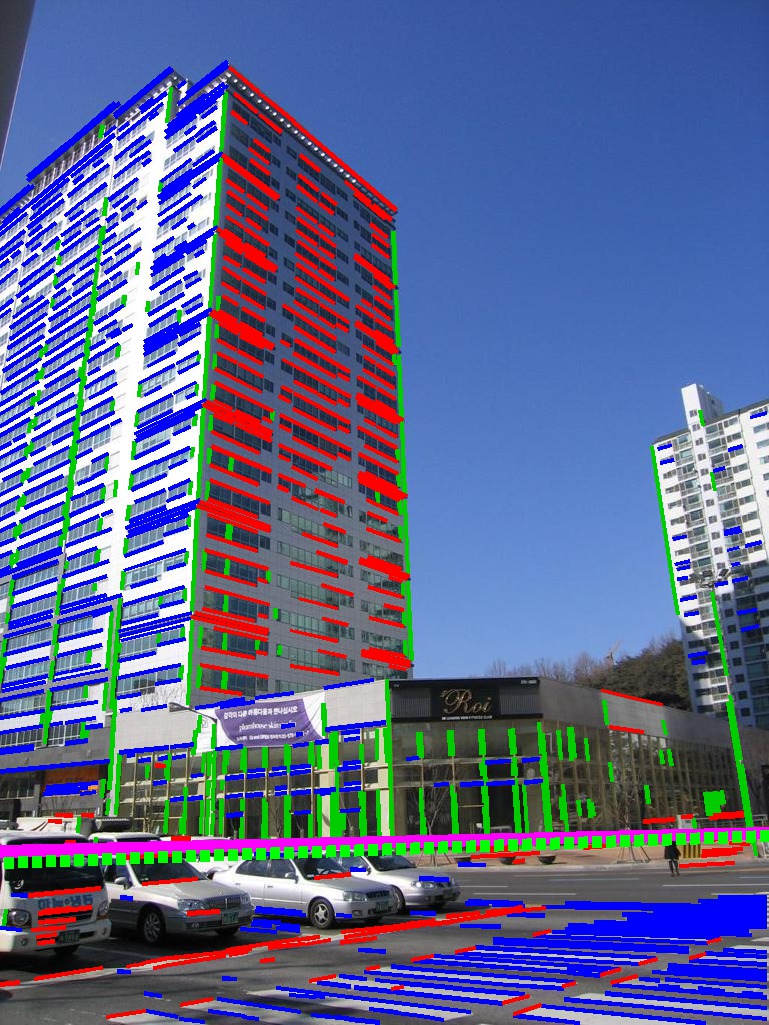}
  \includegraphics[width=0.1540\linewidth]{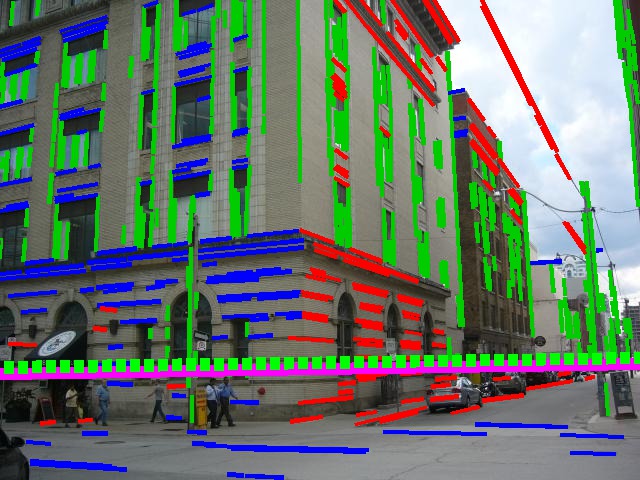}
  \includegraphics[width=0.1540\linewidth]{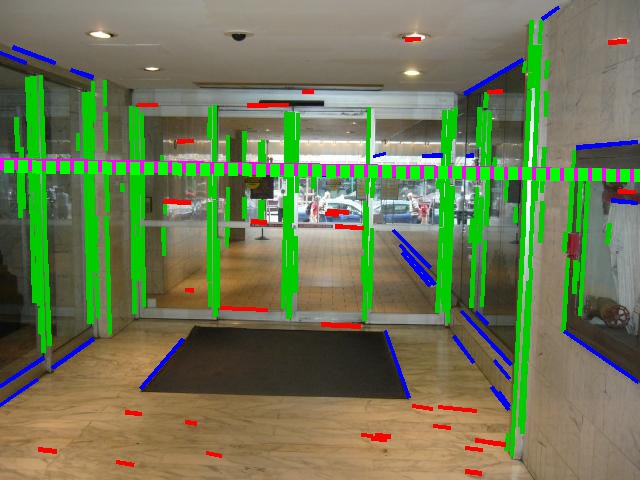}
  \includegraphics[width=0.1731\linewidth]{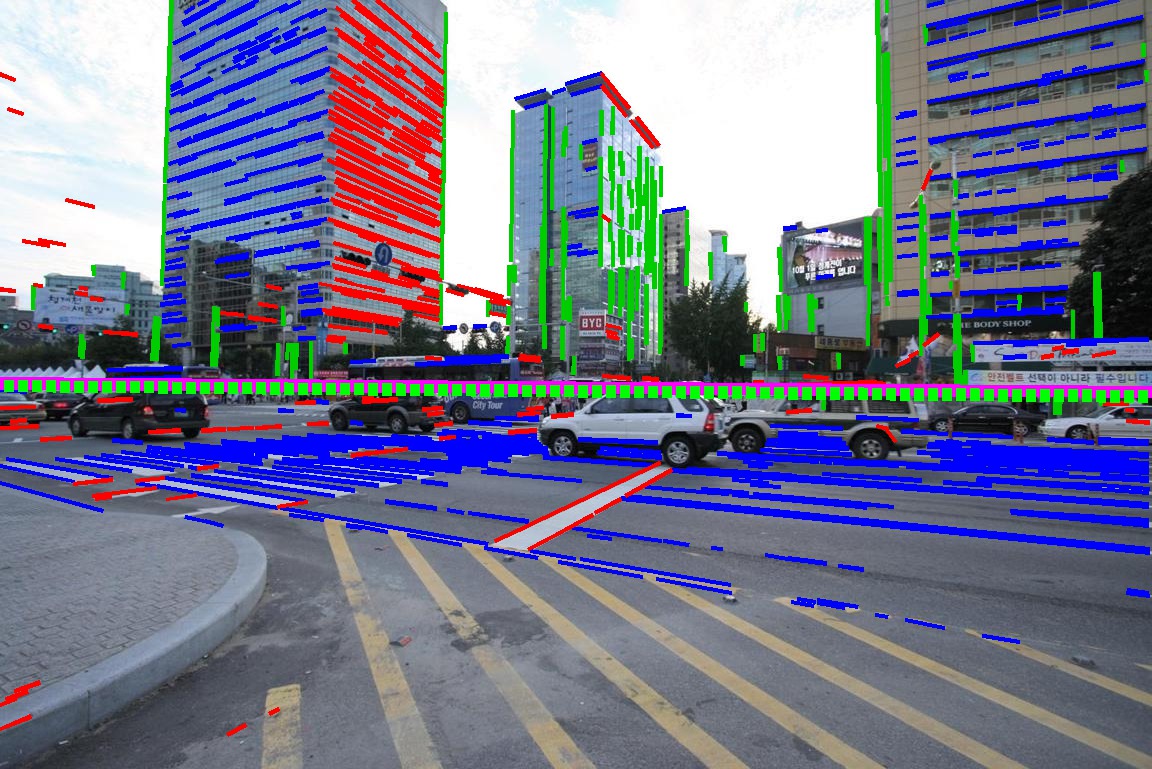}
  \includegraphics[width=0.1540\linewidth]{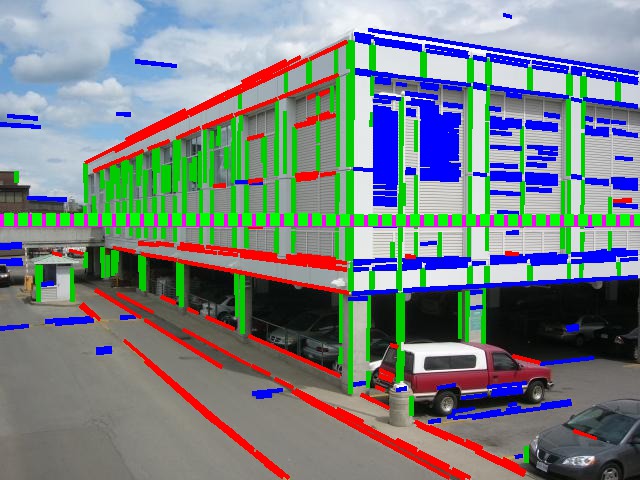}
  \includegraphics[width=0.0867\linewidth]{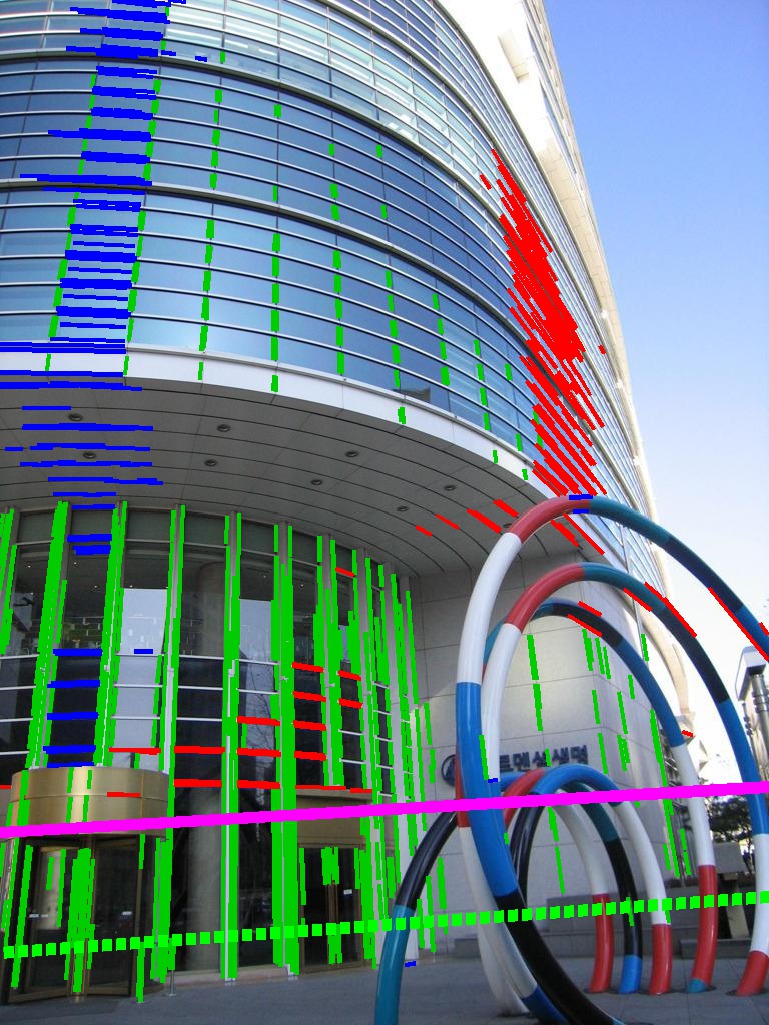}
  \includegraphics[width=0.1375\linewidth]{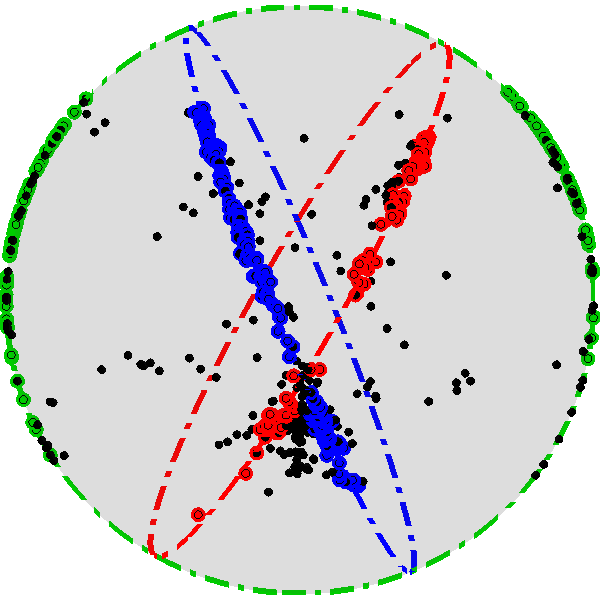}
  \includegraphics[width=0.1375\linewidth]{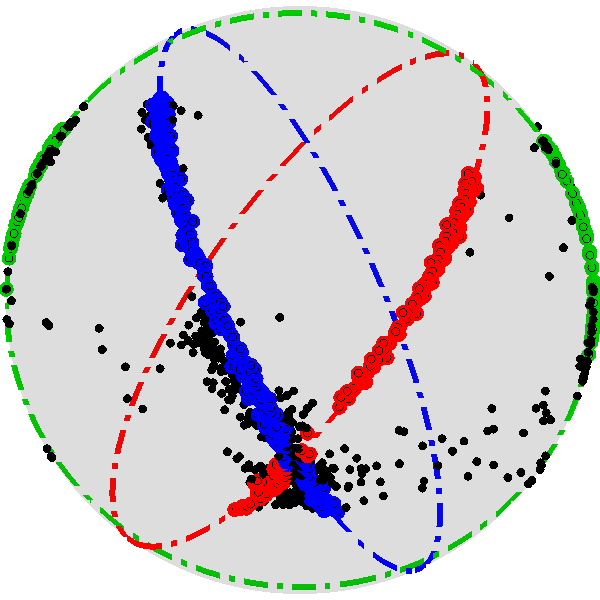}
  \includegraphics[width=0.1375\linewidth]{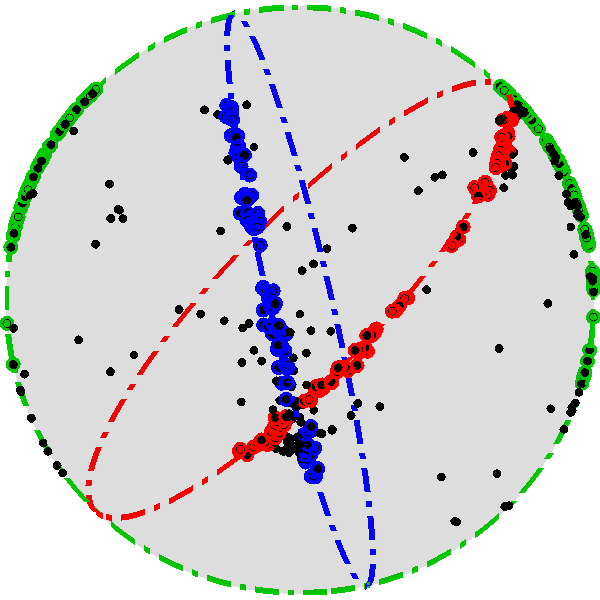}
  \includegraphics[width=0.1375\linewidth]{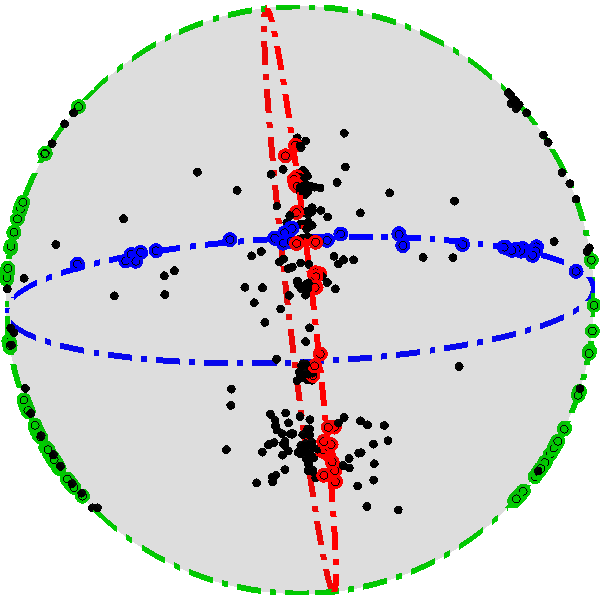}
  \includegraphics[width=0.1375\linewidth]{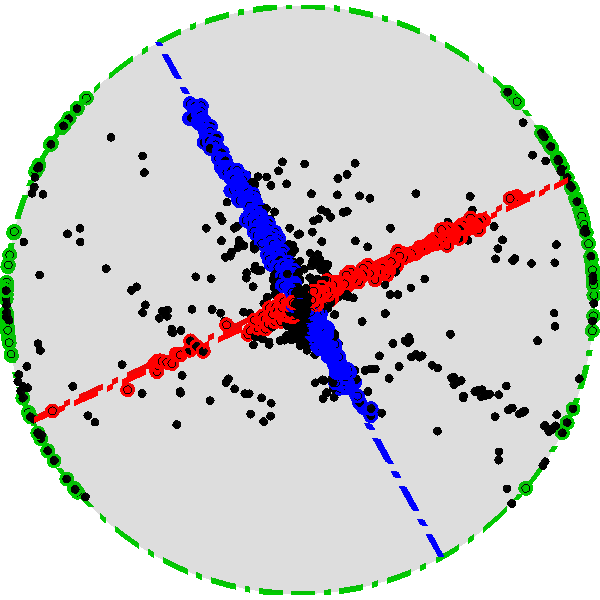}
  \includegraphics[width=0.1375\linewidth]{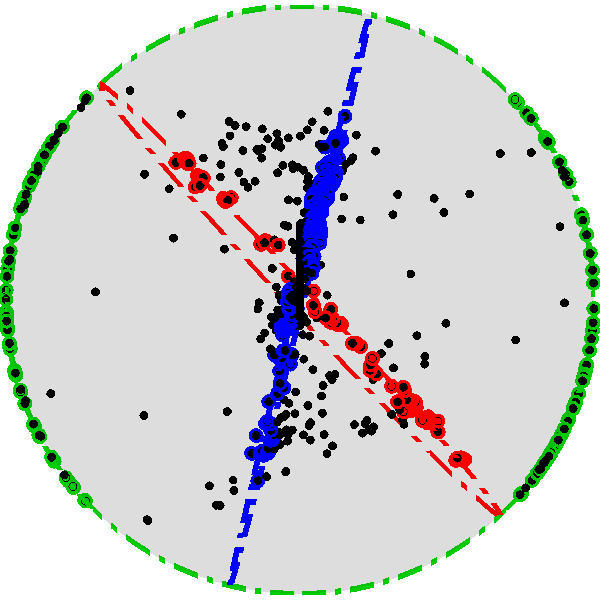}
  \includegraphics[width=0.1375\linewidth]{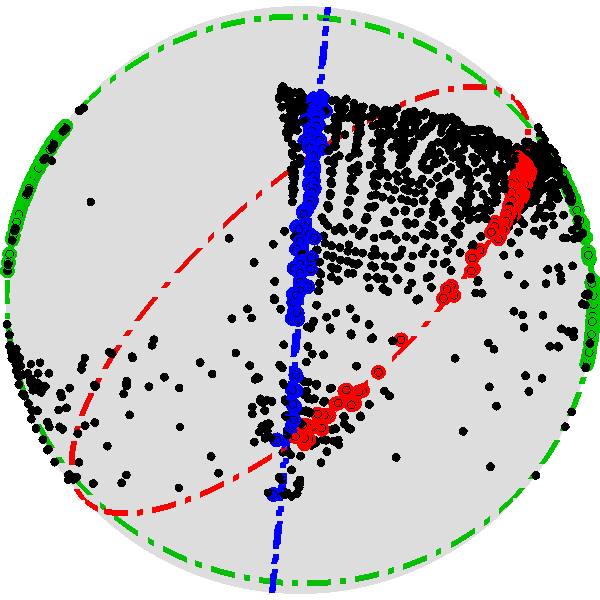}
  \caption{Example results produced by our method.  (rows 1 and 3)
    Line segments color coded based on the most consistent VP, the
    ground-truth (green dash), and detected horizon lines (magenta).
    For clarity only the top two horizontal VPs are shown.  (rows 2
    and 4) The line segments (dots) and their VPs (rings) represented
    in homogeneous coordinates. (last column) Two failure cases of our
    method, caused by irregularly shaped objects (bottom) and short
    edges (top).}
  \label{fig:examples}
\end{figure*}

\subsection{Failure Cases}

We highlight two representative failure cases in the last column of
\figref{examples}.  The top image fails due to the propagation of
measurement errors from the short line segments. The bottom image is
challenging because the curved structures lead to indistinct VPs.
Despite this, global context helps our method produce plausible
results, while other methods (\eg, \cite{geoparser2010}) fail
dramatically.



\section{Conclusion}

We presented a novel vanishing point detection algorithm that obtains
state-of-the-art performance on three benchmark datasets. The main
innovation in our method is the use of global image context to sample
possible horizon lines, followed by a novel discrete-continuous
procedure to score each horizon line by choosing the optimal vanishing
points for the line. Our method is both more accurate and more
efficient than the previous state-of-the-art algorithm, requiring no
parameter tuning for a new testing dataset, which is common in other
methods.


\ifcvprfinal
\section*{Acknowledgements}

We gratefully acknowledge the support of DARPA (contract CSSG
D11AP00255). The U.S.\ Government is authorized to reproduce and
distribute reprints for Governmental purposes notwithstanding any
copyright annotation thereon. Disclaimer: The views and conclusions
contained herein are those of the authors and should not be
interpreted as necessarily representing the official policies or
endorsements, either expressed or implied, of DARPA or the U.S.\
Government.


\fi

{\small
  \bibliographystyle{ieee}
  \bibliography{references}
}

\end{document}